\documentclass[12pt]{article}
\usepackage{graphicx}

\usepackage{etoolbox}
\newtoggle{colt}
\togglefalse{colt}

\usepackage[margin=0.75in]{geometry}
\usepackage{natbib}
\usepackage{boxedminipage}
\usepackage{multirow,nicefrac}
\usepackage{makecell,upgreek}
\usepackage{footnote}
\usepackage{longtable}
\usepackage{tablefootnote}
\usepackage[T1]{fontenc}
\usepackage{verbatim} 
\usepackage[utf8]{inputenc}

\usepackage{booktabs}   
\usepackage{float}

\iftoggle{colt}{}{
\usepackage[dvipsnames]{xcolor}
}

\iftoggle{colt}{}{
\usepackage{amsthm}

\usepackage[breaklinks=true]{hyperref}
\hypersetup{
	colorlinks=true,
	linkcolor=blue,
	citecolor=blue,
	urlcolor=blue
}
}

\iftoggle{colt}{}{
\usepackage{amsfonts,amssymb,mathrsfs}
\usepackage{mathtools}
\usepackage[capitalise,nameinlink]{cleveref}

\usepackage{appendix}

\theoremstyle{plain}
	\newtheorem{theorem}{Theorem}

	\newtheorem{lemma}{Lemma}

	\newtheorem{proposition}{Proposition}

	\theoremstyle{definition}

	\newtheorem{definition}{Definition}

	\newtheorem{remark}{Remark}

}

\iftoggle{colt}{}{
\usepackage{algorithm}
\usepackage[noend]{algpseudocode}
}

\usepackage{autonum}

\def\ddefloop#1{\ifx\ddefloop#1\else\ddef{#1}\expandafter\ddefloop\fi}
\def\ddef#1{\expandafter\def\csname bb#1\endcsname{\ensuremath{\mathbb{#1}}}}
\ddefloop ABCDEFGHIJKLMNOPQRSTUVWXYZ\ddefloop

\def\ddefloop#1{\ifx\ddefloop#1\else\ddef{#1}\expandafter\ddefloop\fi}
\def\ddef#1{\expandafter\def\csname frak#1\endcsname{\ensuremath{\mathfrak{#1}}}}
\ddefloop ABCDEFGHIJKLMNOPQRSTUVWXYZ\ddefloop

\def\ddefloop#1{\ifx\ddefloop#1\else\ddef{#1}\expandafter\ddefloop\fi}
\def\ddef#1{\expandafter\def\csname fr#1\endcsname{\ensuremath{\mathfrak{#1}}}}
\ddefloop ABCDEFGHIJKLMNOPQRSTUVWXYZ\ddefloop

\def\ddefloop#1{\ifx\ddefloop#1\else\ddef{#1}\expandafter\ddefloop\fi}
\def\ddef#1{\expandafter\def\csname eul#1\endcsname{\ensuremath{\EuScript{#1}}}}
\ddefloop ABCDEFGHIJKLMNOPQRSTUVWXYZ\ddefloop

\def\ddefloop#1{\ifx\ddefloop#1\else\ddef{#1}\expandafter\ddefloop\fi}
\def\ddef#1{\expandafter\def\csname scr#1\endcsname{\ensuremath{\mathscr{#1}}}}
\ddefloop ABCDEFGHIJKLMNOPQRSTUVWXYZ\ddefloop

\def\ddefloop#1{\ifx\ddefloop#1\else\ddef{#1}\expandafter\ddefloop\fi}
\def\ddef#1{\expandafter\def\csname b#1\endcsname{\ensuremath{\mathbf{#1}}}}
\ddefloop ABCDEGHIJKLMNOPQRSTUVWXYZ\ddefloop

\def\ddefloop#1{\ifx\ddefloop#1\else\ddef{#1}\expandafter\ddefloop\fi}
\def\ddef#1{\expandafter\def\csname bhat#1\endcsname{\ensuremath{\hat{\mathbf{#1}}}}}
\ddefloop ABCDEFGHIJKLMNOPQRSTUVWXYZ\ddefloop

\def\ddefloop#1{\ifx\ddefloop#1\else\ddef{#1}\expandafter\ddefloop\fi}
\def\ddef#1{\expandafter\def\csname btil#1\endcsname{\ensuremath{\tilde{\mathbf{#1}}}}}
\ddefloop ABCDEFGHIJKLMNOPQRSTUVWXYZ\ddefloop

\def\ddefloop#1{\ifx\ddefloop#1\else\ddef{#1}\expandafter\ddefloop\fi}
\def\ddef#1{\expandafter\def\csname bst#1\endcsname{\ensuremath{\mathbf{#1}^\star}}}
\ddefloop ABCDEFGHIJKLMNOPQRSTUVWXYZ\ddefloop

\def\ddefloop#1{\ifx\ddefloop#1\else\ddef{#1}\expandafter\ddefloop\fi}
\def\ddef#1{\expandafter\def\csname bst#1\endcsname{\ensuremath{\mathbf{#1}^\star}}}
\ddefloop abcdeghijklmnopqrstuvwxyz\ddefloop

\def\ddefloop#1{\ifx\ddefloop#1\else\ddef{#1}\expandafter\ddefloop\fi}
\def\ddef#1{\expandafter\def\csname bhat#1\endcsname{\ensuremath{\hat{\mathbf{#1}}}}}
\ddefloop abcdefghijklmnopqrstuvwxyz\ddefloop

\def\ddefloop#1{\ifx\ddefloop#1\else\ddef{#1}\expandafter\ddefloop\fi}
\def\ddef#1{\expandafter\def\csname b#1\endcsname{\ensuremath{\mathbf{#1}}}}
\ddefloop abcdeghijklnopqrstuvwxyz\ddefloop

\def\ddefloop#1{\ifx\ddefloop#1\else\ddef{#1}\expandafter\ddefloop\fi}
\def\ddef#1{\expandafter\def\csname barb#1\endcsname{\ensuremath{\bar{\mathbf{#1}}}}}
\ddefloop abcdefghijklmnopqrstuvwxyz\ddefloop

\def\ddef#1{\expandafter\def\csname c#1\endcsname{\ensuremath{\mathcal{#1}}}}
\ddefloop ABCDEFGHIJKLMNOPQRSTUVWXYZ\ddefloop
\def\ddef#1{\expandafter\def\csname h#1\endcsname{\ensuremath{\widehat{#1}}}}
\ddefloop ABCDEFGHIJKLMNOPQRSTUVWXYZ\ddefloop
\def\ddef#1{\expandafter\def\csname hc#1\endcsname{\ensuremath{\widehat{\mathcal{#1}}}}}
\ddefloop ABCDEFGHIJKLMNOPQRSTUVWXYZ\ddefloop
\def\ddef#1{\expandafter\def\csname t#1\endcsname{\ensuremath{\widetilde{#1}}}}
\ddefloop ABCDEFGHIJKLMNOPQRSTUVWXYZ\ddefloop
\def\ddef#1{\expandafter\def\csname tc#1\endcsname{\ensuremath{\widetilde{\mathcal{#1}}}}}
\ddefloop ABCDEFGHIJKLMNOPQRSTUVWXYZ\ddefloop

\newcommand{\norm}[1]{\left\|#1\right\|}
\newcommand{\abs}[1]{\left|#1\right|}
\newcommand{\inprod}[2]{\left\langle #1, #2 \right\rangle}

\newcommand{\rr}{\mathbb{R}}
\newcommand{\ee}{\mathbb{E}}
\newcommand{\pp}{\mathbb{P}}

\newcommand{\algfont}[1]{\mathsf{#1}}
\newcommand{\erm}{\algfont{ERM}}

\newcommand{\fhat}{\widehat{f}}
\newcommand{\fstar}{f^\star}

\newcommand{\rad}{\mathfrak{R}}

\DeclareMathOperator*{\argmin}{argmin}
\newcommand{\reg}{\mathrm{Reg}}

\newcommand{\deltatil}{\widetilde{\delta}}
\newcommand{\Yhat}{\widehat{Y}}

\newcommand{\Otil}{\widetilde{O}}
\newcommand{\Omegatil}{\widetilde{\Omega}}
\newcommand{\ptil}{\widetilde{p}}
\newcommand{\Thetatil}{\widetilde{\Theta}}
\newcommand{\err}{\mathrm{Err}}

\DeclareMathOperator*{\argmax}{\mathrm{argmax}}

\usepackage{mathrsfs}

\newcommand{\unif}{\mathrm{Unif}}

\newcommand{\vc}{\mathsf{vc}}
\newcommand{\comp}{\mathrm{comp}}
\renewcommand{\epsilon}{\varepsilon}
\newcommand{\chibar}{\overline{\chi}}
\newcommand{\sign}{\mathrm{sign}}
\newcommand{\polylog}{\mathrm{polylog}}
\newcommand{\radoff}{\mathfrak{R}^{\mathrm{off}}}

\usepackage{algorithm}
\usepackage[]{algpseudocode}

\usepackage[]{color-edits}
\addauthor{ab}{red}
\addauthor{as}{cyan}
\addauthor{sr}{magenta}

\usepackage{natbib}

\title{On the Performance of Empirical Risk Minimization with Smoothed Data}

\usepackage{authblk}
\author[1]{Adam Block}
\author[1]{Alexander Rakhlin}
\author[2]{Abhishek Shetty}
\affil[1]{MIT}
\affil[2]{University of California, Berkeley}

\date{}

\begin{document}

\maketitle

\begin{abstract}
    In order to circumvent statistical and computational hardness results in sequential decision-making, recent work has considered smoothed online learning, where the distribution of data at each time is assumed to have bounded likeliehood ratio with respect to a base measure when conditioned on the history.  
While previous works have demonstrated the benefits of smoothness, they have either assumed that the base measure is known to the learner or have presented computationally inefficient algorithms applying only in special cases. 
This work investigates the more general setting where the base measure is \emph{unknown} to the learner, focusing in particular on the performance of Empirical Risk Minimization (ERM) with square loss when the data are well-specified and smooth.  
We show that in this setting, ERM is able to achieve sublinear error whenever a class is learnable with iid data; in particular, ERM achieves error scaling as $\Otil( \sqrt{\comp(\cF)\cdot T} )$, where $\comp(\cF)$ is the statistical complexity of learning $\cF$ with iid data.  
In so doing, we prove a novel norm comparison bound for smoothed data that comprises the first sharp norm comparison for dependent data applying to arbitrary, nonlinear function classes.
We complement these results with a lower bound indicating that our analysis of ERM is essentially tight, establishing a separation in the performance of ERM between smoothed and iid data.
\end{abstract}

\section{Introduction}\label{sec:intro}

A natural approach to statistical learning is Empirical Risk Minimization (ERM), which, given a function class, returns a hypothesis minimizing the empirical loss on collected data.  When the data are independent and identically distributed (iid), strong guarantees for the performance of ERM are known, and it is statistically optimal in certain cases \citep{birge1993rates, yang1999information, kur2023performance}.
Unfortunately, many learning applications require weaker assumptions on the data generation process than independence.
For this reason, there has been interest in online learning (see e.g. \citep{cesa2006prediction}), a setting where data points $X_t$ arrive one at a time and the learner must predict $\Yhat_t$ before observing $Y_t$, with the goal of minimizing the \emph{regret} with respect to the best hypothesis in hindsight in some class of hypotheses $\cF$ after $T$ rounds; critically, in this setting, no assumptions are made on the data. 
Due to this generality, however, there are many simple settings where statistical \citep{littlestone1988learning,ben2009agnostic} or computational \citep{hazan2016computational} lower bounds preclude learning.

To address these shortcomings, recent work has considered the setting of \emph{smoothed online learning} \citep{rakhlin2011online,haghtalab2020smoothed,haghtalab2022smoothed,haghtalab2022oracle,block2022smoothed,bhatt2023smoothed,block2023oracle,block2023smoothed,block2022efficient,block2023sample}, where the existence of some base measure $\mu$ is posited with the property that, for some parameter $\sigma$ governing the difficulty of the data, the law of $X_t$ conditioned on the history has density bounded by $\sigma^{-1}$ with respect to $\mu$.  In this paper, we consider the performance of ERM when the data are smooth and well-specified, i.e., there exists some $\fstar \in \cF$ such that $\ee\left[ Y_t | X_t \right] = \fstar(X_t)$ for all $t$.  In addition to being an interesting regime in its own right, the ability to learn well-specified data has immediate application to contextual and structured bandits \citep{foster2020beyond, foster2021statistical}.  We show that, in contradistinction to the worst-case data regime where even simple function classes such as thresholds on the unit interval are not learnable \citep{ben2009agnostic}, \emph{ERM is capable of learning whenever the covariates are smooth and the outcomes are well-specified}.

In more detail, we show that when the data $(X_t, Y_t)$ are $\sigma$-smooth, $\ee[Y_t | X_t] = \fstar(X_t)$, and $\fhat_t$ is the ERM on the data collected up to time $t-1$, then
\begin{align}\label{eq:main_result}
    \ee\left[ \sum_{t=  1}^T \left( \fhat_t(X_t) - \fstar(X_t) \right)^2 \right] \lesssim \polylog(T) \cdot \sigma^{-1} \cdot \sqrt{\comp(\cF) \cdot T}.
\end{align}
The proof of \eqref{eq:main_result} rests on three main ingredients.  The first is a decoupling inequality that allows us to control the error of ERM on the observed data sequence $X_t$ by the error of ERM on a conditionally independent (tangent) data sequence $X_t'$:
\begin{align}\label{eq:decoupling_intro}
    \ee\left[ \sum_{t = 1}^T \left( \fhat_t(X_t) - \fstar(X_t) \right)^2  \right] &\lesssim \polylog(T) \cdot \sigma^{-1} \cdot \sqrt{T \cdot \ee\left[ \sum_{t = 1}^T \frac{1}{t} \cdot \sum_{s = 1}^{t-1} \left( \fhat_t(X_s') - \fstar(X_s') \right)^2 \right]}.
\end{align}
Such an inequality as above is useful because, in contradistinction to the iid setting, the distribution of the point on which the ERM $\fhat_t$ is being evaluated can be quite different from the distribution of the data $X_t$; \eqref{eq:decoupling_intro} replaces this distribution shift with error on the independent sequence $X_t'$.  The second ingredient is a novel uniform deviation result that implies sharp control of the population norm by the empirical norm uniformly over a bounded function class $\cG: \cX \to [0,1]$ whenever the data are smooth:
\begin{align}\label{eq:norm_comparison_intro}
    \ee\left[ \sup_{g \in \cG} \sum_{t = 1}^T g(X_t') - 2 \cdot g(X_t)  \right] \lesssim \comp(\cG) \cdot \log\left( \frac T\sigma \right).
\end{align}
In the well-studied setting of iid data  \citep{bousquet2002concentration,mendelson2015learning,rakhlin2017empirical,mendelson2021extending}, analogues of \eqref{eq:norm_comparison_intro} allow us to pass from fixed- to random-design regression, controlling 
\begin{align}
    \norm{\fhat - \fstar}_{L^2(P)}^2 \lesssim \norm{\fhat - \fstar}_{n}^2 + \delta_n^2
\end{align}
for some small $\delta_n > 0$, where $X_1, \dots, X_n \sim P$ are iid and $\norm{\cdot}_n$ is the $L^2$ norm on the empirical measure.  Thus, our approach can be viewed as a generalization of this technique to smoothed data. In particular, \eqref{eq:norm_comparison_intro} allows the right hand side of \eqref{eq:decoupling_intro} to be replaced with the error of ERM on the actual data sequence $X_t$; we conclude by applying a symmetrization technique motivated by the Will's functional \citep{mourtada2023universal} to control error of $\fhat_t$ on the $X_{1:t-1}$.

We note that, as the horizon $T$ tends to infinity, the average error of ERM in \eqref{eq:main_result} vanishes whenever a function class is learnable with iid data.  On the other hand, were the data truly iid, we would expect the cumulative error to grow as $O(\log(T))$ as opposed to the polynomial growth above.  
Surprisingly, we find that our analysis is essentially tight, meaning that for a VC class, ERM must suffer error $\Omega(\sqrt{\vc(\cF) \cdot T})$ in the smoothed setting, even under the stronger assumption of realizability, presenting a significant gap between smoothed and iid data.

Previous work has established that the difficulty of learning under smoothed data with potentially adversarial labels $Y_t$ matches that of iid data, i.e., there exist algorithms whose regret scales like $\Otil\left( \sqrt{\comp(\cF) \cdot T \cdot \log(1/\sigma)} \right)$ \citep{haghtalab2022smoothed,block2022smoothed}, where $\comp(\cF)$ is the statistical complexity of $\cF$ (such as $\vc(\cF)$ or Rademacher complexity).
Furthermore, past work has introduced algorithms that are \emph{efficient} with respect to calls to a black-box ERM oracle and attain regret scaling as $\Otil\left( \sqrt{\comp(\cF) \cdot T} \cdot \sigma^{-1/4} \right)$ \citep{haghtalab2022oracle,block2022smoothed}\footnote{The polynomial separation in $\sigma$ between inefficient and efficient algorithms is provably necessary for proper algorithms. For improper algorithms, this remains an interesting open question.}.  For both of these results, however, the base measure $\mu$ is assumed to be known to the learner in the sense that the learner may efficiently sample from $\mu$.
While this access to $\mu$ is reasonable in many cases (see \cite{block2023smoothed,block2023oracle} and references therein), it is desirable to develop algorithms that do not require any knowledge of the base measure\footnote{As first observed in \citet{block2022smoothed} and generalized in \citet{wu2023online}, when the base measure $\mu$ is unknown, logarithmic dependence on $\sigma$ is impossible, even for computationally inefficient algorithms.}.
As ERM itself does not depend on $\mu$, our work comprises the first example of an (oracle-)efficient algorithm for learning with smoothed data when the base measure is unknown.

We now summarize the main contributions of our paper.
\begin{enumerate}
    \item In \Cref{thm:main}, we show that ERM is capable of learning whenever the data are smoothed and well-specified, further justifying its application even in the absence of the strong assumption of iid data.  In the course of the argument, we state and prove \Cref{lem:self_bounded}, which is a deterministic self-bounding result that may see wider use in the future.
    \item In \Cref{thm:tight_norm_comparison}, we prove a novel norm comparison result for smoothed data comprising the first sharp norm comparison for dependent data applying to arbitrary, nonlinear function classes.
    \item In \Cref{thm:erm_lowerbound}, we demonstrate that our analysis of ERM with smoothed data is tight in the sense that ERM must suffer error $\Omega(\sqrt{\vc(\cF) \cdot T})$ in the smoothed setting, even under the stronger assumption of realizability, presenting a significant gap between smoothed and iid data.
\end{enumerate}
Finally, in \Cref{app:small_ball}, we present \Cref{thm:small_ball}, which is a stronger norm comparison result that can be proved under a natural anti-concentration condition.  In particular, we demonstrate that under this condition, the population norm according to any smooothed distribution can be bounded in expectation by the empirical norm on smoothed data.

\section{Notation and Preliminaries}\label{sec:prelims}

In this section we formalize the problem of smoothed online learning with an unknown base measure as well as introduce the prerequisite notions of function class complexity and assorted analytic constructions that we use throughout the paper.

\subsection{Problem Formulation and Smoothness}
To begin, we define the central condition of our work, smoothness.
\begin{definition}
    Let $\cX$ be a set and $\mu \in \Delta(\cX)$ be a probability distribution over $\cX$.  We say that a measure $p \in \Delta(\cX)$ is $\sigma$-smooth with respect to $\mu$ if $\norm{\frac{d p}{d \mu}}_\infty \leq \sigma^{-1}$, where $\norm{\cdot}_\infty$ is the essential supremum.  Given a sequence of data $X_1, \dots, X_T \in \cX$ adapted to a filtration $(\cH_t)_{t\geq0}$, we say that the data are $\sigma$-smooth with respect to $\mu$ if for all $t \in [T]$, the law of $X_t | \cH_{t-1} \sim p_t$ and $p_t$ are $\sigma$-smooth with respect to $\mu$ almost surely.
\end{definition}
We remark that the requirement that the Radon-Nikodim derivative is bounded can be substantially relaxed to an assumption that $p$ lies in an $f$-divergence ball around $\mu$ \citep{block2023sample}; for the sake of simplicity, we consider only the original definition of smoothness.

In this work, we are concerned with the problem of online supervised learning with square loss.  In particular, we let $\cF: \cX \to [-1,1]$ be a function class and suppose that at each time $t$, the learner chooses an estimator $\fhat_t \in \cF$ before seeing an $X_t \in \cX$ and $Y_t \in \rr$.  In particular, we are interested in the well-specified setting, which we now define.
\begin{definition}\label{def:well_specified}
    Let $(X_1, Y_1), \dots, (X_T, Y_T) \in \cX \times \rr$ be a sequence of data adapted to a filtration $(\cH_t)_{t\geq 0}$.  We say that the data are well-specified with respect to a function class $\cF$ if there exists a function $\fstar\in\cF$, measurable with respect to $\cH_0$ such that for all $t \in [T]$, $\ee\left[ Y_t | \cH_{t-1}, X_t \right] = \fstar(X_t)$.  Furthermore, we say that the data are subGaussian if $Y_t = \fstar(X_t) + \eta_t$ where $\eta_t|\cH_{t-1}$ is a mean-zero subGaussian random variable with variance proxy $\nu^2$.
\end{definition}
The goal of the learner is to predict $\fstar$ as well as possible, i.e. to minimize the estimation error:
\begin{align}\label{eq:error_def}
    \err_T = \sum_{t = 1}^T (\fhat_t(X_t) - \fstar(X_t))^2.
\end{align}
We remark that in general online learning, where no assumption of well-specification is made, it is often more common to study regret  $\sum_{t = 1}^T (\fhat_t(X_t) - Y_t)^2 - \inf_{f \in \cF} \sum_{t = 1}^T (f(X_t) - Y_t)^2$.
While regret is a formally stronger guarantee than the estimation error, the latter is a more natural notion in the well-specified case and is sufficient for applications such as contextual bandits \citep{foster2020beyond,foster2021instance} and reinforcement learning \citep{foster2021statistical,foster2023tight}. 
In the special case of realizable data, where $Y_t = \fstar(X_t)$ for all $t \in [T]$, the notions coincide and thus control of error lead to control of regret.  In particular, when $\cF$ is binary valued and the data are realizable, the cumulative error is precisely the number of mistakes the learner makes over the course of $T$ rounds.

A natural algorithm to handle well-specified data is \emph{Empirical Risk Minimization} (ERM), where at time $t$, the learner chooses
\begin{align}\label{eq:ftl_def}
    \fhat_t \in \argmin_{f \in \cF} \sum_{s = 1}^{t-1} \left( f(X_s) - Y_s \right)^2,
\end{align}
the minimizer of the empirical error on the data seen thus far.  While for many function classes the act of finding $\fhat_t$ can be computationally intractable, motivated by empirical heuristics \citep{goodfellow2016deep}, it is standard in much of online learning to treat the ERM as an oracle that the learner can call efficiently \citep{kalai2005efficient,hazan2016computational,block2022smoothed,haghtalab2022oracle}, as it ensures that the computational difficulty of online learning is not significantly worse than that of offline learning.  

\subsection{Measures of Complexity of a Function Class}
Our error bounds are stated in terms of notions of complexity of the function class $\cF$.  In the course of the paper, we primarily consider the Will's functional of $\cF$ \citep{mourtada2023universal}:
\begin{definition}\label{def:wills}
    Let $\cF: \cX \to \rr$ be a function class, fix $Z_1, \dots, Z_m \in \cX$, and let $\xi_1,\ldots,x_m$ be independent standard Gaussian random variables.  Define the Will's functional of $\cF$ on $Z_1, \dots, Z_m$ to be
    \begin{align}
        W_m(\cF) = \ee_\xi\left[ \exp\left( \sup_{f \in \cF} \sum_{i = 1}^m \xi_i \cdot f(Z_i) - \frac 12 \cdot f(Z_i)^2 \right) \right],
    \end{align}
    where $\ee_\xi\left[ \cdot \right]$ denotes expectation with respect to the $\xi_i$'s, and the dependence on the $Z_i$ is implicit.
\end{definition}
Comparisons between the Will's functional and other standard notions of complexity like Rademacher complexity and covering numbers are well-understood \citep{mourtada2023universal} and we detail some of these connections in \Cref{app:wills}; of particular note is the fact that $\log W_m(\cF) = o(m)$ is necessary and sufficient to ensure statistical learnability with polynomially many samples when the data are iid.  A more standard measure of function class complexity is the Rademacher complexity:
\begin{definition}\label{def:rademacher_complexity}
    Let $\cF: \cX \to [-1,1]$ denote a function class, $\mu \in \Delta(\cX)$ a measure, and $Z_1, \dots, Z_m \sim \mu$ be independent samples from $\mu$.  We define the Rademacher complexity of $\cF$ to be
    \begin{align}
        \rad_m(\cF) = \ee\left[ \sup_{f \in \cF}  \sum_{i = 1}^m \epsilon_i \cdot f(Z_i) \right],
    \end{align}
    where the $\epsilon_i$ are independent Rademacher random variables.
\end{definition}
The Rademacher complexity characterizes the difficulty of distribution-free statistical learning when data are iid and its connections to other standard notions of complexity like the VC dimension \citep{vapnik1999overview} are well-known \citep{van2014probability,wainwright2019high}.  In particular, \citet[Proposition 3.2]{mourtada2023universal} implies\footnote{Technically this result uses the related notion of Gaussian complexity as an upper bound; however, Gaussian complexity is well-known to upper bound Rademacher complexity up to a logarithmic factor \citep{van2014probability}.} that $\log W_m(\cF) \lesssim \rad_m(\cF) \log(m)$ for all $m \in \bbN$.  It is often convenient to instantiate our bounds in the parametric setting for the sake of concreteness.  Thus, we also consider the notion of VC dimension:
\begin{definition}\label{def:vc_dim}
    Let $\cF: \cX \to \left\{ \pm 1 \right\}$ be a function class.  We say that $\cF$ \emph{shatters} points $x_1, \dots, x_d \in \cX$  $\cF$ if for all $\epsilon_{1:d} \in \left\{ \pm 1 \right\}^d$, there is some $f_\epsilon \in \cF$ such that $f_\epsilon(x_i) = \epsilon_i$ for all $i \in [d]$.  We define the VC dimension of $\cF$, denoted by $\vc(\cF)$ to be the maximal $d$ such that there exist points $x_1, \dots, x_d \in \cX$ shattered by $\cF$.
\end{definition}

We note that $\log W_m(\cF) \lesssim \vc(\cF) \cdot \log(m)$ for all $m \in \bbN$ and if $\cF$ is finite, then $\log W_m(\cF) \lesssim \log(\abs{\cF})$ (cf. \Cref{app:wills}).

\subsection{Additional Prerequisites}
A common technique in our analysis is the following coupling lemma, proved in \citet{haghtalab2022smoothed} for discrete $\cX$ and \citet{block2022smoothed} in general.
\begin{lemma}\label{lem:coupling}
    Let $X_1, \dots, X_T$ be $\sigma$-smooth with respect to $\mu$.  Then for all $k \in \bbN$, there exists a coupling of $X_1, \dots, X_T$ with random variables $\left\{ Z_{t,j} | t \in [T], \, j \in [k] \right\}$ such that the $Z_{t,j} \sim \mu$ are independent and there is an event $\cE$ with probability at least $1 - T e^{-\sigma k}$ on which it holds that $X_t \in \left\{ Z_{t,j} | j \in [k] \right\} $ for all $t \in [T]$.
\end{lemma}
This lemma amounts to the key difference between smooth and worst-case data and is one of the reasons that sample acess to the base measure $\mu$ is a central technique in earlier work on smoothed online learning \citep{block2022smoothed,haghtalab2022oracle}.  We use this result purely for analysis as we do not assume that $\mu$ is known to the learner.

Finally, an essential feature of our analysis is a decoupling inequality that disentangles the dependence of the $f_t$ on the data $X_t$.  To this end, we define the following notion of a tangent sequence \citep{decoupling1999}:
\begin{definition}\label{def:tangent_sequence}
    Let $X_t \in \cX$ denote a sequence of random variables adapted to a filtration $(\cH_t)_{t\geq0}$.  We say that a sequence $X_1', \dots, X_T' \in \cX$ is a tangent sequence if for all $t \in [T]$, $X_t$ and $X_t'$ are independent and identically distributed conditioned on $\cH_{t-1}$. 
\end{definition}
Tangent sequences are in general useful for decoupling arguments and have been used  to prove sequential uniform laws of large numbers \citep{rakhlin2015sequential} among many other applications.

\paragraph{Notation.}  We denote by $[T]$ the set $\left\{ 1, \dots, T \right\}$.  
We reserve $\pp$ and $\ee$ to signify probability and expectation when the measure is clear from context.  We let $\Delta(\cX)$ denote the space of distributions on a set $\cX$ and for a measure $\mu \in \Delta(\cX)$, we let $\norm{\cdot}_\mu$ denote the $L^2(\mu)$ norm, i.e. $\norm{f}_\mu = \sqrt{\ee_{Z \sim \mu}[f(Z)^2]}$; in particular, for $t \in [T]$, we let $\norm{\cdot}_t$ denote the empirical norm on the data $X_1, \dots, X_t$.  We use $O(\cdot)$ notation to hide universal constants and $\Otil(\cdot)$ notation to hide polylogarithmic factors.  
\section{Main Results}\label{sec:well_specified}

The main result of this paper is the following bound on the performance of $\fhat_t$:
\begin{theorem}\label{thm:main}
    Let $\cF: \cX \to [-1,1]$ be a function class.  Suppose that $(X_t, Y_t)_{t \in [T]}$ is a sequence of well-specified data such that the $X_t$ are $\sigma$-smooth with respect to some measure $\mu$ and suppose that the $Y_t$ are conditionally $\nu^2$-subGaussian for some $\nu \geq 0$.  Suppose the learner chooses $\fhat_t$ as in \eqref{eq:ftl_def}.  Then\footnote{Note that the lack of a quadratic dependence on $\nu$ does not imply a lack of homogeneity, because the scale of the problem is set by the uniform bound on $\cF$.},
    \begin{align}\label{eq:main}
        \ee\left[ \err_T \right] \leq \frac{20 \log^3(T)}{\sigma} \cdot \sqrt{ T (1+\nu)\left( 1 +  \log \ee_\mu\left[ W_{2T\log(T)/\sigma}(256 \cdot \cF) \right]\right)}.
    \end{align}
\end{theorem}
While \Cref{thm:main} applies to arbitrarily complex, even nonparametric function classes, the clearest instantiation of the result is for parametric function classes where $\log W_m(\cF) = O\left( d \cdot \log(m) \right)$ for fixed $d > 0$ and all $m$, for example when $\vc(\cF) \leq d$.  In this case, we see that the performance of $\fhat_t$ is controlled by $\Otil\left( \sigma^{-1} \sqrt{d \cdot T}\right)$.  While the $\sqrt{T}$ rate is a far cry from the $O(\log(T))$ error guarantees possible when the data are independent, we will see in \Cref{sec:lower_bound} that such logarithmic rates are not in general possible to achieve by ERM with smoothed data.  

As another example, we oberve that whenever $\log W_T(\cF) = o(T / \polylog(T))$, the upper bound \eqref{eq:main} is also $o(T)$.  Due to the fact that sublinear growth in $\log W_m(\cF)$ characterizes learnability in the fixed-design setting \citep{mourtada2023universal}, we see that \Cref{thm:main} is essentially \emph{qualitatively} tight, in the sense that it implies that $\fhat_t$ yields vanishing error whenever the function class $\cF$ is statistically learnable with polynomial rates.  In the special case where the data are \emph{realizable}, error and the more typical notion of regret coincide and thus  \Cref{thm:main} implies a mistake bound.  In particular, for binary-valued $\cF$, we obtain a nontrivial mistake bound for smoothed data simply by playing ERM, which stands in marked contrast to the case of adversarial data.
\begin{remark}
    One immediate application of \Cref{thm:main} is to contextual bandits \citep{lattimore2020bandit}, which is a common partial information setting in sequential decision making.  In this regime, the learner receives \emph{contexts} $X_t$ one at a time before choosing an action $A_t \in [K]$ and observing the reward $Y_t$ depending on the context and action chosen.  Critically, the learner does not observe the counterfactual rewards for actions not chosen.  It is often assumed that the average reward function $\ee\left[ Y_t | X_t, A_t \right] = \fstar(X_t, A_t)$ for some $\fstar \in \cF$ \citep{foster2020beyond,foster2021instance,foster2021efficient} and the goal of the learner is to minimize the regret with respect to the policy induced by $\fstar$.
    By applying the reduction of \citet{foster2020beyond}, we see immediately that if the contexts are smooth, then running \citet[Algorithm 1]{foster2020beyond} with $\fhat_t$ from \eqref{eq:ftl_def} yields a no-regret guarantee for contextual bandits, whenever the function class $\cF$ is statistically learnable.  For example, if $\cF$ is parametric in the sense that $\log W_T(\cF) \lesssim d \cdot \log(T)$, the resulting regret is $\Otil\left(\sigma^{-1/2} K^{3/2} d^{1/4} T^{3/4} \right)$, which is the first nontrivial regret bound for an oracle-efficient algorithm for contextual bandits when the contexts are smooth with respect to an unknown base measure.
\end{remark}
We sketch the proof of \Cref{thm:main} in some detail in the subsequent section, but we highlight one key step here, which may be of independent interest.  In particular, we provide a sharp norm comparison result for smoothed data, comparing the `population norm' of a function on a tangent sequence to the `empirical norm' of the function on the actual data.  This result is the following:
\begin{theorem}\label{thm:tight_norm_comparison}
    Let $\cF: \cX \to [-1,1]$ be a bounded function class and let $X_1, \dots, X_T$ be a sequence of data $\sigma$-smooth with respect to some base measure $\mu \in \Delta(\cX)$.  Then it holds for any $c > 0$ that
    \begin{align}
        \ee\left[ \sup_{f \in \cF} \sum_{t = 1}^T f^2(X_t') - (1+2c) \cdot f^2(X_t)  \right] \leq \sqrt{\frac \pi 2} \cdot \frac{(1+c)^2}{c} \cdot \log \ee_{\mu}\left[ W_{2 T\log(T)/\sigma}\left( \frac{4c}{1+c} \cdot \cF \right) \right] + 4(1+c),
    \end{align}
    where $X_t'$ is a tangent sequence and $W_{t}$ is the Will's functional conditioned on data independently sampled from $\mu$, defined in \Cref{def:wills}.
\end{theorem}
The benefit of \Cref{thm:tight_norm_comparison} in comparison to a more standard uniform deviations approach is that it allows for sharper dependence on the horizon by allowing a small constant factor in front of the empirical norm.  Such a tradeoff is common in norm comparison results for iid data \citep{bousquet2002concentration,mendelson2015learning,mendelson2021extending} and for linear functions of dependent data \citep{simchowitz2018learning,tu2022learning,ziemann2022learning}, but \Cref{thm:tight_norm_comparison} is the first example for dependent data and arbitrary function classes in the literature.  

To understand the power of the new norm comparison, consider the previously known approach using uniform deviations.  Indeed, by combining \citet[Theorem 3]{rakhlin2011online} with \citet[Lemma 17]{block2022smoothed} it is immediate that
\begin{align}\label{eq:uniform_deviations_bad}
    \ee\left[ \sup_{f \in \cF} \sum_{t = 1}^T f^2(X_t') - f^2(X_t) \right] \lesssim \rad_{T\log(T)/\sigma}(\cF^2).
\end{align}
For the sake of completeness, we prove this result as \Cref{lem:uniform_deviations} in \Cref{app:lemmata}.  The problem with applying uniform deviations is that even in the case where $\cF$ is finite, the best bound that \eqref{eq:uniform_deviations_bad} can hope to yield scales like $\Otil(\sqrt{\log(\abs{\cF}) \cdot T})$; this is because $\rad_m(\cF^2)$ is \emph{not} meaningfully smaller than $\rad_m(\cF)$.  Letting $\ptil_T = \frac 1T \sum_{t = 1}^T p_t$, we see that \eqref{eq:uniform_deviations_bad} then implies that for any $f\in \cF$ depending arbitrarily on the data $X_1, \dots, X_T$, we have the bound
\begin{align}
    \ee\left[ \norm{f}_{\ptil_T}^2 \right]  \lesssim \ee\left[ \norm{f}_T^2 \right] + \sqrt{\frac{\log(\abs{\cF})}{T}}.
\end{align}
On the other hand, taking $c$ to be some small constant, \Cref{thm:tight_norm_comparison} yields a bound
\begin{align}
    \ee\left[ \norm{f}_{\ptil_T}^2 \right]  \lesssim \ee\left[ \norm{f}_T^2 \right] + \frac{\log(\abs{\cF})}{T},
\end{align}
which is a significant improvement.  We emphasize that by \citet[Proposition 3.2]{mourtada2023universal}, the logarithm of the Will's functional is never more than a logarithmic factor larger than the Rademacher complexity, and so \Cref{thm:tight_norm_comparison} always yields at least as strong control as \eqref{eq:uniform_deviations_bad} up to a logarithmic factor.

\section{Analysis Techniques}\label{sec:analysis_techniques}
While we defer a detailed proof of \Cref{thm:main,thm:tight_norm_comparison} to \Cref{app:main_proof,app:sharp_norm_comparison} respectively, we here sketch the main idea of the proofs.  In contradistinction to analyzing ERM with iid data, where it suffices to prove a uniform deviation bound to relate predictions on independent test samples to those on training data, for smoothed data there is a \emph{distribution shift} problem where even the distribution on which $\fhat_t$ is being evaluated (that of the next point $p_t$) may not match the distributions of the training data $X_1, \dots, X_{t-1}$.  Thus the first step in the proof of \Cref{thm:main} is to apply a decoupling result, which leverages smoothness of the data to remove this distribution shift.  Unfortunately, upon applying this decoupling, we are left with controlling the performance of ERM on a \emph{tangent sequence}.  It is here that we apply \Cref{thm:tight_norm_comparison} to bound this error by the performance of ERM on the actual data sequence $X_1,\dots, X_T$.  Finally, we will apply a subtle symmetrization argument to conclude the proof.  We begin this section by presenting a more detailed sketch of the preceding summarized argument.  We then sketch the proof of \Cref{thm:tight_norm_comparison}.

\subsection{Proof Sketch of Theorem \ref{thm:main}}

As described above, the proof of \Cref{thm:main} can be broken into three steps: decoupling, norm comparison, and symmetrization.  The first step is to remove the distribution shift with the following decoupling inequality:
\begin{align}\label{eq:decoupling_analysis}
    \ee\left[ \sum_{t = 1}^T \left( \fhat_t(X_t) - \fstar(X_t) \right)^2  \right] &\lesssim \frac{\polylog(T)}{\sigma} \cdot \sqrt{T \cdot \ee\left[ \sum_{t = 1}^T \frac{1}{t} \cdot \sum_{s = 1}^{t-1} \left( f_t(X_s') - \fstar(X_s') \right)^2 \right]}.
\end{align}
The second step is to apply \Cref{thm:tight_norm_comparison} and reduce the problem to bounding $\ee\left[ \norm{\fhat_t - \fstar}_{t-1}^2 \right]$.  The final step is to show that
\begin{align}\label{eq:symmetrization_analysis}
    \ee\left[ \norm{\fhat_t - \fstar}_{t-1}^2 \right] \lesssim \polylog(T) \cdot \log W_{T\log(T)/\sigma}(\cF).
\end{align}
Combining \eqref{eq:decoupling_analysis}, \Cref{thm:tight_norm_comparison}, and \eqref{eq:symmetrization_analysis} then yields the desired result.  We now expand on the first and third steps of the proof and defer discussion of the proof of \Cref{thm:tight_norm_comparison} to the sequel.

\paragraph{Decoupling.}  We begin with the following intermediate result applying to deterministic sequences of bounded real numbers, which we use to prove our decoupling.
\begin{lemma}\label{lem:self_bounded}
    Let $(a_t)_{t \in \bbN}$ denote a sequence of real numbers such that $a_0 = 1$ and $0 \leq a_t \leq 1$ for all $t > 0$.  For $K > 0$ and $t \in \bbN$, let
    \begin{align}
        B_t(a, K) = \left\{ s < t \bigg| a_s \geq \frac Ks \cdot \sum_{u < s} a_u \right\}.
    \end{align}
    Then for any $\epsilon \in (0,1)$, it holds that $\abs{B_T(a, K)} \leq \epsilon T$ for all $K \geq \frac{2 \log(T)}{\epsilon}$.
\end{lemma}
Essentially, the lemma bounds the number of `surprises' a bounded, nonnegative sequence can have, where a `surprise' is a time where an element is significantly larger than the empirical average of the sequence up to that point.  We observe that \Cref{lem:self_bounded} gives more fine-grained control than the more standard so-called ``elliptic potential'' results such as \citet[Lemma 4]{xie2022role}; indeed, whereas these results control the average size of a `surprise,' they yield no control on their number.  On the other hand, \citet[Lemma 4]{xie2022role} follows readily from \Cref{lem:self_bounded}.  While we defer a proof of \Cref{lem:self_bounded} to \Cref{app:self_bounded}, we remark that the proof follows by modifying the sequence $(a_t)$ to a new sequence $(b_t)$ such that $\abs{B_T(b, K)} \geq \abs{B_T(a, K)}$ and the new sequence $(b_t)$ posesses a particular structure amenable to analysis.

The relevance of \Cref{lem:self_bounded} is that it allows us to decouple the estimates $\fhat_t$ from the data $X_t$ by applying the result to the sequence of $a_t = \sigma \cdot \frac{d p_t}{d \mu}(Z)$ for $Z \sim \mu$, where $X_t | \cH_{t-1} \sim p_t$.  In particular, we have the following direct corollary:
\begin{lemma}\label{lem:decoupling}
    Let $(X_t) \subset \cX$ be a sequence of random variables and let $g_t: \cX \to [0,1]$ be a sequence of random functions adapted to a filtration $(\cH_t)_{t\geq0}$ such that $g_t$ is $\cH_{t-1}$-measurable and $X_t | (\cH_{t-1}, g_{t})$ is $\sigma$-smooth with respect to some measure $\mu$.  Let $X_s'$ be a tangent sequence as in \Cref{def:tangent_sequence}.  Then it holds that
    \begin{align}\label{eq:decoupling}
        \ee\left[ \sum_{t = 1}^T g_t(X_t) \right] \leq \frac{\log^2(T)}{\sigma} \cdot \sqrt{2 T \cdot \ee\left[ \sum_{t = 1}^T \frac{1}{t} \cdot \sum_{s = 1}^{t-1} g_t(X_s') \right]}.
    \end{align}
\end{lemma}
\Cref{lem:decoupling} is proved by balancing $\epsilon$ in the application of \Cref{lem:self_bounded}.  In the special case that $\sigma = 1$, however, we see that $a_t = 1$ for all $t$ and thus $\abs{B_T(a, K)} = 0$ for all $K > 1$ and so no balance is needed.  In this case we obtain that the left hand side of \eqref{eq:decoupling} is bounded by $O\left(\log(T) + \ee\left[ \sum_{t=  1}^T g_t(X_t') \right]  \right)$, which is optimal up to constants and the additional logarithmic term, because if $X_t$ are iid, then $\ee[g_t(X_t)] = \ee[g_t(X_t')]$ for all $t$.  Thus we see that our approach to analyzing ERM when specialized to iid data recovers the standard rates up to logarithmic terms and constants.  We further note that a similar result could be achieved through applying the techniques of \citet{xie2022role}, although the proof of an analogous statement in that work is significantly more involved.  We apply \Cref{lem:decoupling} by letting $g_t = (\fhat_t - \fstar)^2$, which yields \eqref{eq:decoupling_analysis}.

\paragraph{Symmetrization.}  The final step in the proof of \Cref{thm:main}, and the only one which requires $\fhat_t$ to be the ERM as opposed to an arbitrary member of $\cF$ depending on $X_1, \dots, X_{t-1}$, is to apply a symmetrization argument to control the estimation error of $\fhat_t$ on the data sequence $X_1, \dots, X_{t-1}$.  We emphasize here that standard symmetrization arguments do not directly apply due to the dependence between the noise $\eta_t$ and the data $X_t$.  Instead, we apply a more subtle symmetrization argument, which takes advantage of the coupling argument of \Cref{lem:coupling}.  In particular, we have the following result:
\begin{lemma}\label{lem:basic_inequality}
    Suppose that $\cF: \cX \to [-1,1]$ is a function class, $\mu \in \Delta(\cX)$, and $X_1, \dots, X_T$ are $\sigma$-smooth with respect to $\mu$.  Suppose further that $Y_t$ are well-specified and $\nu^2$-subGaussian with respect to $\cF$.  Let $\fhat_T$ be $\erm$ on $\cF$ with respect to the data.  For any $k \in \bbN$, it holds that 
    \begin{align}
        \ee\left[ \norm{\fhat_T - \fstar}_{T-1}^2 \right] \leq \frac{64}{T} \nu \cdot \sqrt{\log\left( T \right)} \cdot \left( \log \ee_{Z_{t,j}}\left[W_{k(T-1)}\left( 256 (\cF - \fstar) \right)  \right] + T e^{-\sigma k} \right),
    \end{align}
\end{lemma}
\iftoggle{colt}{\paragraph{Proof sketch.}}{\begin{proof}[Proof sketch.]}

    The full proof of \Cref{lem:basic_inequality} is in \Cref{app:wills_functional} but we provide a sketch here.  By applying elementary computation, we obtain that
    \begin{align}\label{eq:basic_inequality}
        \ee\left[(T-1) \cdot \norm{\fhat_T -\fstar}_{T-1}^2 \right] \leq \ee\left[ \sup_{f \in \cF} 8 \cdot \sum_{t = 1}^{T-1} \eta_t \cdot (f(X_t) - \fstar(X_t)) - \frac 12 \cdot (f(X_t) - \fstar(X_t))^2 \right].
    \end{align}
    We then apply the coupling argument from \Cref{lem:coupling} to separate the right hand side of \eqref{eq:basic_inequality} into a high probability event $\cE$ where $X_t \in \left\{ Z_{t,j} \right\}$ for $Z_{t,j} \sim \mu$ independent and the low probability complement.  On the high probability event, we then symmetrize, observe that the $\eta$ can be dropped by passing to their worst-case absolute value, and apply Jensen's inequality to upper bound the right hand side of \eqref{eq:basic_inequality} by
    \begin{align}
        \frac 1\lambda \log \ee\left[  \exp\left( \bbI[\cE] \cdot \lambda \cdot \sup_{f \in \cF} 8 \cdot \sum_{t = 1}^{T-1} \xi_t \cdot (f(X_t) - \fstar(X_t)) - \frac 12 \cdot (f(X_t) - \fstar(X_t))^2  \right) \right] + \frac 1T,
    \end{align} 
    where the $\xi_t$ are independent standard Gaussians and $\lambda$ is a carefully chosen constant.  Finally, we conclude the proof by using the coupling as well as the monotonicity of the Will's functional proved in \Cref{lem:monotonicity} to replace the $X_t$ with $Z_{t,j}$. \iftoggle{colt}{\jmlrQED}{\end{proof}}
    \begin{remark}\label{rmk:will_needed}
        We emphasize that passing to the Will's functional before applying the coupling is essential.  Indeed, the key fact that we use about the Will's functional is that $W_m(\cF) \leq W_{m+1}(\cF)$ for all $m$, which allows us to replace $X_t$ (which has a complicated dependence on $\xi_1,\dots,\xi_{t-1}$) with the independent $Z_{t,j}$.  This monotonicity property does not hold for the right hand side of \eqref{eq:basic_inequality} and so we cannot directly apply the coupling to it.
    \end{remark}

    \Cref{lem:basic_inequality} says that if the data are smooth and the labels are well-specified, then the expected performance of the $\erm$ $\fhat_t$ on the historical data $X_1, \dots, X_T$ is controlled by the Will's functional, which is, in turn, well-behaved when $\cF$ is a simple class.  Combining \Cref{lem:basic_inequality} with the preceding argument then concludes the proof of \Cref{thm:main}.

\subsection{Proof Sketch of Theorem \ref{thm:tight_norm_comparison}}
While we defer a detailed proof to \Cref{app:sharp_norm_comparison}, we provide a brief sketch here.  The proof proceeds by adapting the \emph{tree of probabilities} construction from \citet{rakhlin2011online} in order to apply symmetrization, and then using a variation of the coupling result, \Cref{lem:coupling} along with Jensen's inequality to pass to the Will's functional on iid data.  In more detail, we first observe, as in the proof of  \citet[Lemma 18]{liang2015learning}, that 
\begin{align}
    \ee\left[ \sup_{f \in \cF} \sum_{t = 1}^T f(X_t')^2 - (1+2c) \cdot f(X_t)^2  \right] &= \ee\left[ \sup_{f \in \cF} \sum_{t = 1}^T (1+c) (f(X_t')^2 -  f(X_t)^2) - c f(X_t')^2 - c f(X_t)^2 \right].\label{eq:norm_comparison_pf_analysis}
\end{align}
and note that the first term in the right hand side is \emph{anti-symmetric} in $(X_t,X_t')$ while the second term is \emph{symmetric}.  We then introduce the tree of probabilities construction from \citet{rakhlin2011online} and construct a measure $\rho$ on a $\cX$-valued complete binary trees $\bx$ such that the right hand side of \eqref{eq:norm_comparison_pf_analysis} is upper bounded by 
\begin{align}\label{eq:norm_comparison_pf_analysis_2}
    2(1+c) \cdot \ee_{\bx \sim \rho}\left[ \sup_{f \in \cF} \sum_{t = 1}^T \xi_t f^2(\bx_t(\xi)) - \frac c{1+c} f^4(\bx_t(\xi)) \right],
\end{align}
where $\bx_t(\epsilon) \sim p_t$ is $\sigma$-smooth.  We then apply a variant (\Cref{lem:tree_coupling}) of the coupling result \Cref{lem:coupling} above to introduce an event $\cE$ with high probability at least $1 - T e^{- \sigma k}$ such that $\bx_t(\epsilon) \in \left\{ Z_{t,j} | j \in [k] \right\}$ for all $t$.  As in \Cref{rmk:will_needed}, we cannot directly apply the coupling as \eqref{eq:norm_comparison_pf_analysis_2} is not necessarily monotone in $T$.  Instead, we apply a similar technique as was done in the proof of \Cref{lem:basic_inequality} and upper bound \eqref{eq:norm_comparison_pf_analysis_2} by
\begin{align}
    2(1+c) \left( \frac 1\lambda \log \ee\left[ \exp\left(\bbI[\cE] \lambda \cdot \sup_{f \in \cF} \sum_{t = 1}^T \xi_t f^2(\bx_t(\xi)) - \frac c{1+c} f^4(\bx_t(\xi))  \right) \right] + T^2 e^{-\sigma k} \right).
\end{align}
Now we may apply the same monotonicity result, \Cref{lem:monotonicity} as in that previous proof to pass to independent data and observe that the resulting expression is just the Will's functional applied to the function class $\cF^2$.  The proof concludes by noting that if $\cF$ is uniformly bounded, then $f \mapsto f^2$ is uniformly Lipschitz and the Will's functional satisfies contraction with respect to Lipschitz functions \citep[Theorem 4.1]{mourtada2023universal}. \iftoggle{colt}{\jmlrQED}{\qed}

\begin{remark}
    Note that the term subtracted from \eqref{eq:norm_comparison_pf_analysis_2} contains $f^4$ instead of $f^2$; the fact that this is an upper bound on the same expression with $f^2$ being subtracted is immediate from the boundedness of $\cF$, but the quartic power is key here in order to pass to the Will's functional.  The analogous result for iid data, \citet[Lemma 18]{liang2015learning} does not require this technique because one can apply a contraction argument that is not available in the more general, smoothed data regime.
\end{remark}

\section{Lower Bound for ERM}\label{sec:lower_bound}
As we have seen above, for parametric classes ERM is able to achieve $\Otil\left( \sigma^{-1} \sqrt{d T} \right)$ error whenever the data are smooth and well-specified.  While this results in an asymptotically no-regret guarantee, the rate is far from the $O(d \cdot \log(T))$ error that is known to be achievable when $\sigma = 1$ and the data are independent and identically distributed \citep{wainwright2019high}.  In this section we demonstrate the surprising fact that ERM is unable to obtain these so-called `fast rates' when the data are merely smooth as opposed to iid.  We emphasize that we do not rule out the possibility of oracle-efficient algorithms achieving these fast rates, but simply demonstrate that the most natural algorithm for learning with smoothed data is not competitive, even in the realizable setting.  This is the content of the following result.
\begin{theorem}\label{thm:erm_lowerbound}
    For any $d \in \bbN$ there exists a function class $\cF$ with $\vc(\cF) = d$ such that for any $0 < \sigma < 1$ and any horizon $T$, there is a $\sigma$-smooth adversary realizable with respect to $\cF$ such that if $\Yhat_t = \fhat_t(X_t)$ is always chosen such that $\fhat_t$ is an ERM in \eqref{eq:ftl_def}, then
    \begin{align}
        \ee\left[ \err_T \right] \geq \frac 12 \cdot \sqrt{d \cdot T \cdot \frac{1 - \sigma^{1/d}}{\sigma^{1/d}}}.
    \end{align}
\end{theorem}
Note that in the special case of $\sigma = 1$, when the data are iid, \Cref{thm:erm_lowerbound} is vacuous, as expected.  On the other hand, for $\sigma \ll 1$, we see that ERM can never hope to do better than $\Omega(\sqrt{d T})$, which is significantly worse than the logarithmic-in-$T$ guarantees from statistical learning.   The proof of \Cref{thm:erm_lowerbound} is deferred to \Cref{app:erm_lowerbound}, but we sketch the construction in the $d = 1$ case here.

\iftoggle{colt}{\paragraph{Proof sketch of \Cref{thm:erm_lowerbound}.}}{\begin{proof}[Proof sketch]}  We consider $\cX = [0,1]$ the unit interval and $\cF$ the class of thresholds, with an adversary that samples $X_t \sim p_t$, where $p_t = \unif([j \epsilon, j \epsilon + \sigma])$, $j$ is the number of mistakes made up to time $t - 1$, and $\epsilon > 0$ is a tuning parameter; we let the $Y_t = 0$ for all $t$, making the data realizable with respect to $\cF$.  The key observation is that we may choose the ERM to predict $1$ as frequently as posible conditioned on fitting all of the data thus far.  Thus, whenever $M_t = \max_{s \leq t} X_s$ increases, this choice of ERM will always predict incorrectly.  In this way the adversary can only force $(1-\sigma) / \epsilon$ mistakes until the unit interval is fully covered, and each mistake happens with probablity $\epsilon / \sigma$ at each time step $t$.  In expectation, then, the number of mistakes made is at least $\min(\epsilon/\sigma T, (1-\sigma)/\epsilon)$.  Balancing $\epsilon$ yields the desired result for $d = 1$; the $d > 1$ case is then just a tensorized version of this construciton. \iftoggle{colt}{\jmlrQED}{\end{proof}}

Combining \Cref{thm:erm_lowerbound} with \Cref{thm:main} we see our analysis of ERM is tight in its dependence on complexity and horizon, i.e., ERM achieves $\Thetatil(\sqrt{\vc(\cF) \cdot T})$ error whenever the data are smooth for $\sigma < 1$.  We leave the interesting question of whether other oracle-efficient algorithms can achieve improved error in this setting as an interesting direction for future research.

\section{Related Work}
In this section, we briefly survey some related work and place our results in the context of recent literature on oracle efficiency in smoothed online learning and norm comparison bounds for population and empirical norms.

\paragraph{Smoothed Online Learning.} Given the statistical and computational intractibility of learning with adversarial data, many recent works have investigated the difficulty of online learning with beyond-worst-case assumptions.  In particular, \cite{rakhlin2011online} presented a general framework for online learning against adversaries that are somehow constrained in each round and characterized the minimax regret through a quantity called the \emph{distribution-dependent sequential Rademacher complexity}.  Following this work, \citet{haghtalab2022smoothed} considered the \emph{smooth} setting and demonstrated that minimax regret of classification can be greatly improved when the data are smooth with respect to a known base measure; these results were later extended to regression in \citet{block2022smoothed} and to more general notions of smoothness in \citet{block2023sample}.  More recently, smoothed online learning has been applied to a variety of settings including sequential probability assignment \citep{bhatt2023smoothed}, learning in auctions \citep{durvasula2023smoothed,cesa2023repeated}, and robotics \citep{block2023oracle,block2023smoothed}.  The case where the base measure is unknown has seen relatively less attention, with \citet{block2022smoothed} observing that guarantees for smoothed online learning with an unknown base measure are necessarily worse than those where $\mu$ is known and \citet{wu2023online} providing statistical bounds in a particular special case.  We emphasize that in all of the above works, the focus has been on general Lipschitz losses, with the squared loss being treated as a special case.  While this suffices for qualitative results with bounded function classes, it is well-known that the additional curvature of the square loss admits faster statistical rates with both iid \citep{birge1993rates,bousquet2002concentration,liang2015learning} and adversarial data \citep{rakhlin2014online}.  Our work demonstrates that, unlike the case of iid data, ERM itself is unable to achieve these faster rates in the smoothed setting.

Beyond the setting of full-information online learning, \cite{xie2022role} analyzed the role of smoothness (termed \emph{coverability}) in online reinforcement learning.  In that work, the authors proved a decoupling result similar to and motivating our \Cref{lem:decoupling}, which forms the starting point of our analysis.
 While \citet{xie2022role} go on to apply this decoupling result to prove guarantees for a computationally inefficient algorithm in RL, we instead focus on its implications to efficient algorithms for online learning.

\paragraph{Oracle Efficiency in Online Learning.}  A major problem in the study of computational efficiency in online learning is the provable hardness of many optimization tasks, which are strictly easier than online learning.
Motivated by efficient algorithms in combinatorial optimization and the empirical success of optimization heuristics in function classes of interest \citep{goodfellow2016deep}, many works have assumed access to an optimization oracle that is efficiently able to minimize an empirical loss function on data over a function class \citep{kalai2005efficient}, with \citet{hazan2016computational} demonstrating the limits thereof.
In the context of smoothed online learning, several works have circumvented the computational lower bounds of \citet{hazan2016computational} with oracle-efficient algorithms applying in variations on the smoothed setting \citep{block2022smoothed,haghtalab2022oracle,block2022efficient,block2023oracle,block2023sample,block2023smoothed}.
To our knowledge, our work is the first to analyze an oracle-efficient algorithm (in fact, ERM itself) for the smoothed online setting when the base measure is unknown.

\paragraph{Population and Empirical Norm Comparisons.} It has long been important in nonparametric statistics and learning theory to understand comparisons between empirical and population norms that hold uniformly over function classes \citep{bousquet2002concentration}.  Of particular note is the `small-ball method' of  \citet{koltchinskii2015bounding,mendelson2015learning,mendelson2021extending}, that introduces an approach to such comparisons relying on anti-concentration that holds for independent data in great generality.  In the case of sequential data, much less is known, with most all work focusing on norm comparison results holding for \emph{linear function classes} \citep{abbasi2011improved,simchowitz2018learning,ziemann2022learning,tu2022learning}.  In this work, we provide the first sharp norm comparison result for general, \emph{nonlinear} function classes that holds whenever the data are smooth and a certain small-ball condition is satisfied.  Most relevant to our work is the approach of \citet{liang2015learning}, which introduces \emph{offset Rademacher complexity} as a tighter form of control for sharp norm comparison.  While we take inspiration from this approach, a direct application of these techniques does not work due to the lack of monotonicity of this measure and our resulting inability to apply the coupling.  Instead, we control the relaxed complexity notion fo the Will's functional, which was extensively explored in \citet{mourtada2023universal}.

\section*{Acknowledgements}
We acknowledge support from ARO through award W911NF-21-1-0328, the Simons Foundation and the NSF through awards DMS-2031883 and DMS-1953181.  In addition, AB acknowledges support from the National Science Foundation Graduate Research Fellowship under Grant No.1122374 and AS acknowledges support from the Apple AI+ML fellowship.

\bibliographystyle{plainnat}
\bibliography{refs}

\begin{thebibliography}{54}
\providecommand{\natexlab}[1]{#1}
\providecommand{\url}[1]{\texttt{#1}}
\expandafter\ifx\csname urlstyle\endcsname\relax
  \providecommand{\doi}[1]{doi: #1}\else
  \providecommand{\doi}{doi: \begingroup \urlstyle{rm}\Url}\fi

\bibitem[Abbasi-Yadkori et~al.(2011)Abbasi-Yadkori, P{\'a}l, and Szepesv{\'a}ri]{abbasi2011improved}
Yasin Abbasi-Yadkori, D{\'a}vid P{\'a}l, and Csaba Szepesv{\'a}ri.
\newblock Improved algorithms for linear stochastic bandits.
\newblock \emph{Advances in neural information processing systems}, 24, 2011.

\bibitem[Azuma(1967)]{azuma1967weighted}
Kazuoki Azuma.
\newblock Weighted sums of certain dependent random variables.
\newblock \emph{Tohoku Mathematical Journal, Second Series}, 19\penalty0 (3):\penalty0 357--367, 1967.

\bibitem[Ben-David et~al.(2009)Ben-David, P{\'a}l, and Shalev-Shwartz]{ben2009agnostic}
Shai Ben-David, D{\'a}vid P{\'a}l, and Shai Shalev-Shwartz.
\newblock Agnostic online learning.
\newblock 2009.

\bibitem[Bhatt et~al.(2023)Bhatt, Haghtalab, and Shetty]{bhatt2023smoothed}
Alankrita Bhatt, Nika Haghtalab, and Abhishek Shetty.
\newblock Smoothed analysis of sequential probability assignment.
\newblock \emph{arXiv preprint arXiv:2303.04845}, 2023.

\bibitem[Birg{\'e} and Massart(1993)]{birge1993rates}
Lucien Birg{\'e} and Pascal Massart.
\newblock Rates of convergence for minimum contrast estimators.
\newblock \emph{Probability Theory and Related Fields}, 97:\penalty0 113--150, 1993.

\bibitem[Block and Polyanskiy(2023)]{block2023sample}
Adam Block and Yury Polyanskiy.
\newblock The sample complexity of approximate rejection sampling with applications to smoothed online learning.
\newblock In Gergely Neu and Lorenzo Rosasco, editors, \emph{Proceedings of Thirty Sixth Conference on Learning Theory}, volume 195 of \emph{Proceedings of Machine Learning Research}, pages 228--273. PMLR, 12--15 Jul 2023.
\newblock URL \url{https://proceedings.mlr.press/v195/block23a.html}.

\bibitem[Block and Simchowitz(2022)]{block2022efficient}
Adam Block and Max Simchowitz.
\newblock Efficient and near-optimal smoothed online learning for generalized linear functions.
\newblock \emph{Advances in Neural Information Processing Systems}, 35:\penalty0 7477--7489, 2022.

\bibitem[Block et~al.(2022)Block, Dagan, Golowich, and Rakhlin]{block2022smoothed}
Adam Block, Yuval Dagan, Noah Golowich, and Alexander Rakhlin.
\newblock Smoothed online learning is as easy as statistical learning.
\newblock In \emph{Conference on Learning Theory}, pages 1716--1786. PMLR, 2022.

\bibitem[Block et~al.(2023{\natexlab{a}})Block, Simchowitz, and Rakhlin]{block2023oracle}
Adam Block, Max Simchowitz, and Alexander Rakhlin.
\newblock Oracle-efficient smoothed online learning for piecewise continuous decision making.
\newblock In Gergely Neu and Lorenzo Rosasco, editors, \emph{Proceedings of Thirty Sixth Conference on Learning Theory}, volume 195 of \emph{Proceedings of Machine Learning Research}, pages 1618--1665. PMLR, 12--15 Jul 2023{\natexlab{a}}.
\newblock URL \url{https://proceedings.mlr.press/v195/block23b.html}.

\bibitem[Block et~al.(2023{\natexlab{b}})Block, Simchowitz, and Tedrake]{block2023smoothed}
Adam Block, Max Simchowitz, and Russ Tedrake.
\newblock Smoothed online learning for prediction in piecewise affine systems.
\newblock In \emph{Advances in Neural Information Processing Systems}. Curran Associates, Inc., 2023{\natexlab{b}}.
\newblock URL \url{https://openreview.net/pdf?id=Izt7rDD7jN}.

\bibitem[Bousquet(2002)]{bousquet2002concentration}
Olivier Bousquet.
\newblock \emph{Concentration inequalities and empirical processes theory applied to the analysis of learning algorithms}.
\newblock PhD thesis, {\'E}cole Polytechnique: Department of Applied Mathematics Paris, France, 2002.

\bibitem[Cesa-Bianchi and Lugosi(2006)]{cesa2006prediction}
Nicolo Cesa-Bianchi and G{\'a}bor Lugosi.
\newblock \emph{Prediction, learning, and games}.
\newblock Cambridge university press, 2006.

\bibitem[Cesa-Bianchi et~al.(2023)Cesa-Bianchi, Cesari, Colomboni, Fusco, and Leonardi]{cesa2023repeated}
Nicol{\`o} Cesa-Bianchi, Tommaso~R Cesari, Roberto Colomboni, Federico Fusco, and Stefano Leonardi.
\newblock Repeated bilateral trade against a smoothed adversary.
\newblock In \emph{The Thirty Sixth Annual Conference on Learning Theory}, pages 1095--1130. PMLR, 2023.

\bibitem[De~la Pena and Gin{\'e}(1999)]{decoupling1999}
Victor De~la Pena and Evarist Gin{\'e}.
\newblock \emph{Decoupling: from dependence to independence}.
\newblock Springer, 1999.

\bibitem[Dudley(1978)]{dudley1978central}
Richard~M Dudley.
\newblock Central limit theorems for empirical measures.
\newblock \emph{The Annals of Probability}, pages 899--929, 1978.

\bibitem[Durvasula et~al.(2023)Durvasula, Haghtalab, and Zampetakis]{durvasula2023smoothed}
Naveen Durvasula, Nika Haghtalab, and Manolis Zampetakis.
\newblock Smoothed analysis of online non-parametric auctions.
\newblock In \emph{Proceedings of the 24th ACM Conference on Economics and Computation}, pages 540--560, 2023.

\bibitem[Foster and Rakhlin(2020)]{foster2020beyond}
Dylan Foster and Alexander Rakhlin.
\newblock Beyond ucb: Optimal and efficient contextual bandits with regression oracles.
\newblock In \emph{International Conference on Machine Learning}, pages 3199--3210. PMLR, 2020.

\bibitem[Foster et~al.(2021{\natexlab{a}})Foster, Rakhlin, Simchi-Levi, and Xu]{foster2021instance}
Dylan Foster, Alexander Rakhlin, David Simchi-Levi, and Yunzong Xu.
\newblock Instance-dependent complexity of contextual bandits and reinforcement learning: A disagreement-based perspective.
\newblock In \emph{Conference on Learning Theory}, pages 2059--2059. PMLR, 2021{\natexlab{a}}.

\bibitem[Foster and Krishnamurthy(2021)]{foster2021efficient}
Dylan~J Foster and Akshay Krishnamurthy.
\newblock Efficient first-order contextual bandits: Prediction, allocation, and triangular discrimination.
\newblock \emph{Advances in Neural Information Processing Systems}, 34:\penalty0 18907--18919, 2021.

\bibitem[Foster et~al.(2021{\natexlab{b}})Foster, Kakade, Qian, and Rakhlin]{foster2021statistical}
Dylan~J Foster, Sham~M Kakade, Jian Qian, and Alexander Rakhlin.
\newblock The statistical complexity of interactive decision making.
\newblock \emph{arXiv preprint arXiv:2112.13487}, 2021{\natexlab{b}}.

\bibitem[Foster et~al.(2023)Foster, Golowich, and Han]{foster2023tight}
Dylan~J Foster, Noah Golowich, and Yanjun Han.
\newblock Tight guarantees for interactive decision making with the decision-estimation coefficient.
\newblock \emph{arXiv preprint arXiv:2301.08215}, 2023.

\bibitem[Goodfellow et~al.(2016)Goodfellow, Bengio, and Courville]{goodfellow2016deep}
Ian Goodfellow, Yoshua Bengio, and Aaron Courville.
\newblock \emph{Deep learning}.
\newblock MIT press, 2016.

\bibitem[Hadwiger(1975)]{hadwiger1975will}
Hugo Hadwiger.
\newblock Das will'sche funktional.
\newblock \emph{Monatshefte f{\"u}r Mathematik}, 79\penalty0 (3):\penalty0 213--221, 1975.

\bibitem[Haghtalab et~al.(2020)Haghtalab, Roughgarden, and Shetty]{haghtalab2020smoothed}
Nika Haghtalab, Tim Roughgarden, and Abhishek Shetty.
\newblock Smoothed analysis of online and differentially private learning.
\newblock \emph{Advances in Neural Information Processing Systems}, 33:\penalty0 9203–9215, 2020.

\bibitem[Haghtalab et~al.(2022{\natexlab{a}})Haghtalab, Han, Shetty, and Yang]{haghtalab2022oracle}
Nika Haghtalab, Yanjun Han, Abhishek Shetty, and Kunhe Yang.
\newblock Oracle-efficient online learning for beyond worst-case adversaries.
\newblock \emph{arXiv e-prints}, pages arXiv--2202, 2022{\natexlab{a}}.

\bibitem[Haghtalab et~al.(2022{\natexlab{b}})Haghtalab, Roughgarden, and Shetty]{haghtalab2022smoothed}
Nika Haghtalab, Tim Roughgarden, and Abhishek Shetty.
\newblock Smoothed analysis with adaptive adversaries.
\newblock In \emph{2021 IEEE 62nd Annual Symposium on Foundations of Computer Science (FOCS)}, pages 942--953. IEEE, 2022{\natexlab{b}}.

\bibitem[Hazan and Koren(2016)]{hazan2016computational}
Elad Hazan and Tomer Koren.
\newblock The computational power of optimization in online learning.
\newblock In \emph{Proceedings of the forty-eighth annual ACM symposium on Theory of Computing}, pages 128--141, 2016.

\bibitem[Kalai and Vempala(2005)]{kalai2005efficient}
Adam Kalai and Santosh Vempala.
\newblock Efficient algorithms for online decision problems.
\newblock \emph{Journal of Computer and System Sciences}, 71\penalty0 (3):\penalty0 291--307, 2005.

\bibitem[Koltchinskii and Mendelson(2015)]{koltchinskii2015bounding}
Vladimir Koltchinskii and Shahar Mendelson.
\newblock Bounding the smallest singular value of a random matrix without concentration.
\newblock \emph{International Mathematics Research Notices}, 2015\penalty0 (23):\penalty0 12991--13008, 2015.

\bibitem[Kur(2023)]{kur2023performance}
Gil Kur.
\newblock \emph{On The Performance Of The Maximum Likelihood Over Large Models}.
\newblock PhD thesis, Massachusetts Institute of Technology, 2023.

\bibitem[Lattimore and Szepesv{\'a}ri(2020)]{lattimore2020bandit}
Tor Lattimore and Csaba Szepesv{\'a}ri.
\newblock \emph{Bandit algorithms}.
\newblock Cambridge University Press, 2020.

\bibitem[Liang et~al.(2015)Liang, Rakhlin, and Sridharan]{liang2015learning}
Tengyuan Liang, Alexander Rakhlin, and Karthik Sridharan.
\newblock Learning with square loss: Localization through offset rademacher complexity.
\newblock In \emph{Conference on Learning Theory}, pages 1260--1285. PMLR, 2015.

\bibitem[Littlestone(1988)]{littlestone1988learning}
Nick Littlestone.
\newblock Learning quickly when irrelevant attributes abound: A new linear-threshold algorithm.
\newblock \emph{Machine learning}, 2:\penalty0 285--318, 1988.

\bibitem[Mendelson(2015)]{mendelson2015learning}
Shahar Mendelson.
\newblock Learning without concentration.
\newblock \emph{Journal of the ACM (JACM)}, 62\penalty0 (3):\penalty0 1--25, 2015.

\bibitem[Mendelson(2021)]{mendelson2021extending}
Shahar Mendelson.
\newblock Extending the scope of the small-ball method.
\newblock \emph{Studia Mathematica}, 256:\penalty0 147--167, 2021.

\bibitem[Mendelson and Vershynin(2003)]{mendelson2003entropy}
Shahar Mendelson and Roman Vershynin.
\newblock Entropy and the combinatorial dimension.
\newblock \emph{Inventiones mathematicae}, 152\penalty0 (1):\penalty0 37--55, 2003.

\bibitem[Mohri et~al.(2018)Mohri, Rostamizadeh, and Talwalkar]{mohri2018foundations}
Mehryar Mohri, Afshin Rostamizadeh, and Ameet Talwalkar.
\newblock \emph{Foundations of machine learning}.
\newblock MIT press, 2018.

\bibitem[Mourtada(2023)]{mourtada2023universal}
Jaouad Mourtada.
\newblock Universal coding, intrinsic volumes, and metric complexity.
\newblock \emph{arXiv preprint arXiv:2303.07279}, 2023.

\bibitem[Rakhlin and Sridharan(2014)]{rakhlin2014online}
Alexander Rakhlin and Karthik Sridharan.
\newblock Online non-parametric regression.
\newblock In \emph{Conference on Learning Theory}, pages 1232--1264. PMLR, 2014.

\bibitem[Rakhlin et~al.(2011)Rakhlin, Sridharan, and Tewari]{rakhlin2011online}
Alexander Rakhlin, Karthik Sridharan, and Ambuj Tewari.
\newblock Online learning: Stochastic, constrained, and smoothed adversaries.
\newblock \emph{Advances in neural information processing systems}, 24, 2011.

\bibitem[Rakhlin et~al.(2015)Rakhlin, Sridharan, and Tewari]{rakhlin2015sequential}
Alexander Rakhlin, Karthik Sridharan, and Ambuj Tewari.
\newblock Sequential complexities and uniform martingale laws of large numbers.
\newblock \emph{Probability theory and related fields}, 161:\penalty0 111--153, 2015.

\bibitem[Rakhlin et~al.(2017)Rakhlin, Sridharan, and Tsybakov]{rakhlin2017empirical}
Alexander Rakhlin, Karthik Sridharan, and Alexandre~B Tsybakov.
\newblock Empirical entropy, minimax regret and minimax risk.
\newblock \emph{Bernoulli}, 23\penalty0 (2):\penalty0 789--824, 2017.

\bibitem[Rudin et~al.(1976)]{rudin1976principles}
Walter Rudin et~al.
\newblock \emph{Principles of mathematical analysis}, volume~3.
\newblock McGraw-hill New York, 1976.

\bibitem[Simchowitz et~al.(2018)Simchowitz, Mania, Tu, Jordan, and Recht]{simchowitz2018learning}
Max Simchowitz, Horia Mania, Stephen Tu, Michael~I Jordan, and Benjamin Recht.
\newblock Learning without mixing: Towards a sharp analysis of linear system identification.
\newblock In \emph{Conference On Learning Theory}, pages 439--473. PMLR, 2018.

\bibitem[Tu et~al.(2022)Tu, Frostig, and Soltanolkotabi]{tu2022learning}
Stephen Tu, Roy Frostig, and Mahdi Soltanolkotabi.
\newblock Learning from many trajectories.
\newblock \emph{arXiv preprint arXiv:2203.17193}, 2022.

\bibitem[Van~Handel(2014)]{van2014probability}
Ramon Van~Handel.
\newblock Probability in high dimension.
\newblock \emph{Lecture Notes (Princeton University)}, 2014.

\bibitem[Vapnik(1999)]{vapnik1999overview}
Vladimir~N Vapnik.
\newblock An overview of statistical learning theory.
\newblock \emph{IEEE transactions on neural networks}, 10\penalty0 (5):\penalty0 988--999, 1999.

\bibitem[Vitale(1996)]{vitale1996wills}
Richard~A Vitale.
\newblock The wills functional and gaussian processes.
\newblock \emph{The Annals of Probability}, 24\penalty0 (4):\penalty0 2172--2178, 1996.

\bibitem[Wainwright(2019)]{wainwright2019high}
Martin~J Wainwright.
\newblock \emph{High-dimensional statistics: A non-asymptotic viewpoint}, volume~48.
\newblock Cambridge university press, 2019.

\bibitem[Wills(1973)]{wills1973gitterpunktanzahl}
J{\"o}rg~M Wills.
\newblock Zur gitterpunktanzahl konvexer mengen.
\newblock \emph{Elemente der Mathematik}, 28:\penalty0 57--63, 1973.

\bibitem[Wu et~al.(2023)Wu, Grama, and Szpankowski]{wu2023online}
Changlong Wu, Ananth Grama, and Wojciech Szpankowski.
\newblock Online learning in dynamically changing environments.
\newblock In Gergely Neu and Lorenzo Rosasco, editors, \emph{Proceedings of Thirty Sixth Conference on Learning Theory}, volume 195 of \emph{Proceedings of Machine Learning Research}, pages 325--358. PMLR, 12--15 Jul 2023.
\newblock URL \url{https://proceedings.mlr.press/v195/wu23a.html}.

\bibitem[Xie et~al.(2022)Xie, Foster, Bai, Jiang, and Kakade]{xie2022role}
Tengyang Xie, Dylan~J Foster, Yu~Bai, Nan Jiang, and Sham~M Kakade.
\newblock The role of coverage in online reinforcement learning.
\newblock In \emph{The Eleventh International Conference on Learning Representations}, 2022.

\bibitem[Yang and Barron(1999)]{yang1999information}
Yuhong Yang and Andrew Barron.
\newblock Information-theoretic determination of minimax rates of convergence.
\newblock \emph{Annals of Statistics}, pages 1564--1599, 1999.

\bibitem[Ziemann and Tu(2022)]{ziemann2022learning}
Ingvar Ziemann and Stephen Tu.
\newblock Learning with little mixing.
\newblock \emph{Advances in Neural Information Processing Systems}, 35:\penalty0 4626--4637, 2022.

\end{thebibliography}

\newpage

\tableofcontents

\appendix

\crefalias{section}{appendix} 

\section{Background on the Will's Functional}\label{app:wills}

The Will's functional is a fundamental quantity originally associated to convex bodies in $\rr^m$ \citep{wills1973gitterpunktanzahl,hadwiger1975will}.  More recently, \citet{mourtada2023universal} extended the definition of the Will's functional to arbitrary subsets of $A \subset \rr^m$ by taking advantage of a Gaussian representation due to \citet{vitale1996wills}.  In that paper, \citet{mourtada2023universal} proves a number of fundamental results about this complexity measure, including contraction and sharp connections with other standard notions.  In this section, we provide a brief overview of the Will's functional's connections to other notions of complexity in learning theory; we defer to the excellent \citet{mourtada2023universal} for a more detailed treatment.  We recall from \Cref{def:wills} that
\begin{align}
    W_m(\cF) = \ee_\xi\left[\exp\left(\sup_{f \in \cF} \sum_{i = 1}^m \xi_i f(Z_i) - \frac 12 \cdot f^2(Z_i)\right)  \right],
\end{align}
where $\xi_i$ are independent standard Gaussians.  One fundamental property is the invariance under translation:
\begin{proposition}[Proposition 3.1.5 in \citet{mourtada2023universal}]\label{prop:translation_invariance}
    Let $\cF$ be a function class $\iota: \rr^m \to \rr^m$ be an affine isometry in that $\iota$ is affine and preserves the Euclidean norm.  Then,
    \begin{align}
        W_m(\iota(\cF)) = W_m(\cF).
    \end{align}
\end{proposition}
A particular case of the above is when $\iota$ is a translation, in which case \Cref{prop:translation_invariance} implies translation invariance.  Another fundamental property is the contraction of the Will's functional under composition with Lipschitz functions:
\begin{proposition}[Theorem 4.1 from \citet{mourtada2023universal}]\label{prop:contraction}
    If $\iota: \rr \to \rr$ is a contraction, in that $\iota$ is $1$-Lipschitz, then $W_m(\iota \circ \cF) \leq W_m(\cF)$.
\end{proposition}
In particular \Cref{prop:contraction} implies monotonicity of the Will's functional, which is a key difference from the related notion of offset Rademacher complexity introduced by \citet{liang2015learning}.  While in our proofs, we require a slightly stronger version of this monotonicity (\Cref{lem:monotonicity}), this property of the Will's functional is what motivates its utility in applying the coupling.

We now recall several results relating the Will's functional to other standard notions of complexity.  The first demonstrates that $W_m(\cF)$ is not much larger than the Rademacher complexity:
\begin{proposition}[Proposition 3.2 from \citet{mourtada2023universal}]\label{prop:gaussian_comp}
    For any class $\cF$, $m \in \bbN$, and dataset $Z_1, \dots, Z_m$, recalling $\rad_m(\cF)$ from \Cref{def:rademacher_complexity}, it holds that
    \begin{align}
        \log W_m(\cF) \lesssim \sqrt{\log(m)} \cdot \rad_m(\cF).
    \end{align}
\end{proposition}
\begin{proof}
    By \citet[Proposition 3.2]{mourtada2023universal}, it holds that
    \begin{align}
        \log W_m(\cF) \leq \ee_\xi \left[ \sup_{f \in \cF} \sum_{i = 1}^m \xi_i f(Z_i) \right],
    \end{align}
    where the uppper bound is the Gaussian complexity.  It is well known that the Gaussian complexity is upper bounded by the Rademacher complexity up to a factor logarithmic in $m$ \citep{van2014probability,wainwright2019high}, and the result follows immediately.
\end{proof}
While $\rad_m(\cF)$ presents an upper bound for the Will's functional, it is not in general tight.  Instead, a lower bound can be found in the \emph{offset Rademacher complexity}.
\begin{proposition}\label{prop:offset_rad}
    Recall from \citet{liang2015learning} that for a function class $\cF$ and data $Z_1, \dots, Z_m$, the offset Rademacher complexity is defined as
    \begin{align}
        \radoff_m(\cF) = \ee_\epsilon \left[ \sup_{f \in \cF} \sum_{i = 1}^m \epsilon_i f(Z_i) - c f^2(Z_i) \right],
    \end{align}
    where $\epsilon_i$ are independent Rademacher random variables and $c > 0$.  Then it holds for any $m$ that $\radoff_m(\cF) \lesssim (2c)^{-1} \cdot \log W_m(2c\cF)$.
\end{proposition}
\begin{proof}
    Letting $\xi_i$ denote independent standard gaussians, we compute by Jensen's inequality for any $\lambda > 0$,
    \begin{align}
        \radoff_m(\cF) &= \ee_\epsilon \left[ \sup_{f \in \cF} \sum_{i = 1}^m \epsilon_i f(Z_i) - c f^2(Z_i) \right] \\
        &=  \ee_\epsilon \left[ \sup_{f \in \cF} \sum_{i = 1}^m \epsilon_i\cdot \frac{\ee[\abs{\xi_i}]}{\ee[\abs{\xi_i}]} f(Z_i) - c f^2(Z_i) \right] \\
        &\leq \sqrt{\frac{\pi}{2}} \cdot \ee_\xi \left[ \sup_{f \in \cF} \sum_{i = 1}^m \xi_i f(Z_i) - c f^2(Z_i) \right] \\
        &\leq \sqrt{\frac \pi 2} \cdot \frac 1\lambda   \cdot \log \ee_\xi \left[\exp\left( \sup_{f \in \cF} \sum_{i = 1}^m \lambda \xi_i f(Z_i) - \lambda c f^2(Z_i) \right)\right].
    \end{align}
    Setting $\lambda = 2c$, we see that
    \begin{align}
        \radoff_m(\cF) \leq \sqrt{\frac \pi 2} \cdot \frac{1}{2c} \log W_m\left( 2 c \cdot \cF \right).
    \end{align}
    The result follows immediately.
\end{proof}
Combining \Cref{prop:gaussian_comp,prop:offset_rad} yields the fact that sublinearity in $m$ of the Will's functional characterizes learnability of a class $\cF$ with polynomially many samples.

Finally, we recall the relationship between the Will's funcitonal and the covering number, which we now define.
\begin{definition}\label{def:covering_number}
    Let $\cF: \cX \to [-1,1]$ be a function class and let $\norm{\cdot}$ denote a norm on $\cF$.  For any scale $\delta > 0$, we say that a set $f_1, \dots, f_m$ of functions is a $\delta$-cover of $\cF$ with respect to $\norm{\cdot}$ if for all $f \in \cF$, there exists $i \in [m]$ such that $\norm{f - f_i} \leq \delta$.  We define the covering number of $\cF$ with respect to $\norm{\cdot}$ to be the minimal size of a $\delta$-cover of $\cF$ with respect to $\norm{\cdot}$ and denote it by $\cN(\cF, \delta, \norm{\cdot})$.
\end{definition}
The covering numbers of many standard function classes are known and this complexity notion and its relationship to Rademacher complexity is well-understood in the context of statistical learning theory \citep{van2014probability,wainwright2019high}.  In particular, if $\vc(\cF) \leq d$, then $\log \cN(\cF, \delta) \lesssim d \log\left( \frac 1\delta \right)$ as $\delta \downarrow 0$ \citep{dudley1978central,mendelson2003entropy}.  The following fundamental result relates this notion to the Will's functional:
\begin{proposition}[Theorem 4.2 from \citet{mourtada2023universal}]\label{prop:covering_number}
    Let $\cF$ be a covering number and define for $r > 0$,
    \begin{align}
        \rad_m(\cF, r) = \sup_{f_0 \in \cF} \rad_m\left( \cF \cap B_r(f_0) \right),
    \end{align}
    where $B_r(f_0)$ is the ball of radius $r$ around $f_0$.  Then it holds that
    \begin{align}
         \inf_{r > 0} \left\{ \rad_m(\cF, r) + \log \cN(\cF, r)\right\} \lesssim \log W_m(\cF) \lesssim \sqrt{\log(m)} \cdot \inf_{r > 0} \left\{ \rad_m(\cF, r) + \log \cN(\cF, r)\right\}.
    \end{align}
\end{proposition}
It follows immediately that if $\cF$ is finite, then $\log W_m(\cF) \lesssim \log(\abs{\cF})$ and if $\cF$ is a VC class, then $\log W_m(\cF) \lesssim d \cdot \log(m)$.
\section{Proof of Lemma \ref{lem:self_bounded}}\label{app:self_bounded}
In this section, we prove \Cref{lem:self_bounded}.  The proof proceeds by modifying the sequence $(a_t)$ to a new sequence $(b_t)$ such that $\abs{B_T(b, K)} \geq \abs{B_T(a, K)}$ and the new sequence $(b_t)$ posesses a particularly easy to analyze structure. 

We first note that it suffices to consider small $K$.
\begin{lemma}\label{lem:small_K}
    Let $(a_t)$ be a sequence as in \Cref{lem:self_bounded}.  If $K > T$, then $B_T(a, K) = \emptyset$.
\end{lemma}
\begin{proof}
    Suppose that $B_T(a, K) \neq \emptyset$ and let $t_1$ be the minimal element of $B_T(a, K)$, whose existence is implied by the nonempty assumption.  Note that
    \begin{align}
        1 \geq a_{t_1} \geq \frac{K}{t_1} \geq \frac{K}{T},
    \end{align}
    where the first inequality follows by construction and the second follows by the fact that $a_t \geq 0$ and the definition of $B_T(a, K)$.  Rearranging concludes the proof.
\end{proof}

We are now ready to prove the lemma.
\begin{proof}[Proof of \Cref{lem:self_bounded}]
    By \Cref{lem:small_K}, it suffices to assume that $K \leq T$.  Furthermore, observing that $\abs{B_T(a, K)}$ is decreasing as $K$ increases, it suffices to prove the claim for $K = \frac{2 \log(T)}{\epsilon}$.  To do this, let $(a_t)$ be a fixed sequence as in the statement of the lemma and fix $K \leq T$.  Let $B_T(a, K) = \left\{ t_1, \dots, t_i \right\}$, i.e., $t_1, \dots, t_i$ are the set of `surprises' where $a_t$ is much larger than expected.  We define a new sequence $(b_t)$ such that $b_0 = 1$, $b_{t_1} = a_{t_1}$, and for $t > 0$,
    \begin{align}
        b_t = \begin{cases}
            0 & t \not\in B_T(a, K) \\
            \frac Kt \cdot\sum_{s < t} b_s & t \in B_T(a, K) \setminus \left\{ t_1 \right\}
        \end{cases}.
    \end{align}
    We prove in \Cref{lem:bt_good} below that that $0 \leq b_t \leq a_t \leq 1$ for all $t \in [T]$ and that $\abs{B_T(b, K)} \geq \abs{B_T(a, K)}$.  Thus it suffices to prove the main claim for $(b_t)$ instead of $(a_t)$.

    To prove the claim for $(b_t)$, we compute:
    \begin{align}
        b_{t_j} &= \frac{K}{t_j} \cdot \sum_{s < t_j} b_s \\
        &= \frac{K}{t_j} \cdot \sum_{s \in B_{t_j}(b, K)} b_s \\
        &= \frac{K}{t_j} \cdot \left( \sum_{s \in B_{t_{j-1}}(a, K)} b_s + b_{t_{j-1}} \right) \\
        &= \frac{K}{t_j} \cdot \left( \frac{t_{j-1}}{K} \cdot b_{t_{j-1}} + b_{t_{j-1}} \right) \\
        &= \frac{K + t_{j-1}}{t_j} \cdot b_{t_{j-1}}.
    \end{align}
    Thus it holds that
    \begin{align}
        1 \geq a_{t_i} \geq b_{t_i} = \frac{K}{t_i} \cdot \prod_{j = 1}^{i-1}\left( 1 + \frac{K}{t_{j-1}} \right).
    \end{align}
    Taking logarithms of both sides and rearranging, we see that
    \begin{align}
        \log\left( \frac{t_i}K \right) &\geq \sum_{j = 1}^{i-1} \log\left( 1 + \frac{K}{t_{j-1}} \right) \geq \sum_{s = T-i}^T \log\left( 1 + \frac Ks \right) \geq i \cdot \log\left( 1 + \frac KT \right),
    \end{align}
    where the second inequality follows by the fact that the $t_j$ are distinct and all at most $T$.  Now we note that as $K \geq 1$ and $t_i \leq T$, it holds that
    \begin{align}
        i \cdot \log\left( 1 + \frac KT \right) \leq \log\left( \frac{t_i}{K} \right) \leq \log(T).
    \end{align}
    Observing that
    \begin{align}
        \log\left( 1 + \frac KT \right) \geq \frac{\frac KT}{1 + \frac KT},
    \end{align}
    we see that
    \begin{align}
        i \leq \log(T) \cdot \frac{1 + \frac KT}{\frac KT} = \frac{T \cdot \log(T)}{K} \left( 1 + \frac KT \right).
    \end{align}
    Letting $K = \frac{2 \log(T)}{\epsilon}$, recalling that $K \leq T$ and thus $1 + \frac KT \leq 2$, and plugging in concludes the proof.
\end{proof}
We now prove the previously deferred result above.
\begin{lemma}\label{lem:bt_good}
    Let $(a_t)$ be a sequence as in \Cref{lem:self_bounded}, $K > 0$ fixed, and 
    \begin{align}
        B_T(a, K) = \left\{ t_1, \dots, t_i \right\} \subset [T].
    \end{align}
    Let $b_0 = 1$, $b_{t_1} = a_{t_1}$, and, for $t > 0$, let
    \begin{align}
        b_t = \begin{cases}
            0 & t \not\in B_T(a, K) \\
            \frac Kt \cdot\sum_{s < t} b_s & t \in B_T(a, K) \setminus \left\{ t_1 \right\}
        \end{cases}.
    \end{align}
    Then $\abs{B_T(b, K)} \geq \abs{B_T(b, K)}$.  Furthermore, for all $t \in [T]$, it holds that $b_t \leq a_t \leq 1$.
\end{lemma}
\begin{proof}
    To see the first point, observe that by construction, $B_T(b, K) \supseteq B_T(a, K)$ and so this claim follows immediately.  To see the second point, we first note that for $t \not\in B_T(a, K)$, we have $b_t = 0 \leq a_t$.  For $t \in B_T(a,K)$, we induct on $j \in [i]$.  Indeed it is clear that $b_{t_1} = a_{t_1}$ and so the claim holds.  Suppose that $b_{t_k} \leq a_{t_k}$ for $k < j \in [i]$.  Then we observe that
    \begin{align}
        b_{t_j} = \frac K{t_j} \cdot \sum_{s < t_j} b_s = \frac K{t_j} \cdot \sum_{k < j} b_{t_k} \leq \frac K{t_j} \cdot \sum_{k < j} a_{t_k} \leq \frac K{t_j} \cdot \sum_{s < t_j} a_s \leq a_{t_j},
    \end{align}
    where the first two equalities follow by construction, the first inequality follows by the inductive hypothesis, the second inequality follows by the fact that $a_t \geq 0$ and the final inequality follows by the fact that $t_j \in B_T(a, K)$.  Thus $b_t \leq a_t \leq 1$ for all $t \in [T]$. 
\end{proof}

\section{Proof of Theorem \ref{thm:main}}\label{app:main_proof}
In this appendix we provide the complete proof of \Cref{thm:main}.  As described in \Cref{sec:well_specified} the proof is split into three parts.  In this appendix, we begin by proving the decoupling inequality in \Cref{lem:decoupling} and then proceed to prove \Cref{lem:basic_inequality} before finally concluding the proof of the main result.  Although we use \Cref{thm:tight_norm_comparison} in the conclusion of the proof of \Cref{thm:main}, we defer its proof to \Cref{app:sharp_norm_comparison}.

\subsection{Proof of Lemma \ref{lem:decoupling}}\label{app:decoupling}

    Let the $g_t$ be as in the statement of the lemma, $p_t$ denote the law of $X_t$ conditioned on the $\sigma$-algebra generated by $(\cH_{t-1}, g_t)$, and $\ptil_t = \frac 1{t} \cdot \sum_{s = 1}^{t-1} \frac{d p_s}{d \mu}$.  
We compute
    \begin{align}
        \ee\left[ \sum_{t = 1}^T g_t(X_t) \right] &= \ee\left[ \sum_{t = 1}^T \ee\left[ g_t(X_t) | g_t, \cH_{t-1} \right] \right] \\
        &= \ee\left[ \sum_{t = 1}^T \ee\left[\frac{d p_t}{d \mu}(Z) g_t(Z)| g_t, \cH_{t-1} \right]\right] \\
        &= \ee_Z \ee_{g_t}\left[ \sum_{t = 1}^T \frac{d p_t}{d \mu}(Z) g_t(Z) \right],
    \end{align}
    where the $Z$ are independent of the $X_1, \dots, X_T$ and the $g_t$ are measurable with respect to $\cH_{t-1}$.  Let $a_t(Z) = \sigma \cdot \frac{d p_t}{d\mu}(Z)$ be a random sequence and observe that by \Cref{lem:self_bounded}, $\abs{B_T(a(Z), K)} \leq \epsilon T$ deterministically whenever $K \geq 2 \log(T)/\epsilon$.  Thus, we see for some fixed $K$ large enough,
    \begin{align}
        \ee_Z \ee_{g_t}\left[ \sum_{t = 1}^T \frac{d p_t}{d \mu}(Z) g_t(Z) \right] &= \ee_Z \ee_{g_t}\left[ \sum_{t = 1}^T \frac{d p_t}{d \mu}(Z) g_t(Z) \bbI\left[ t \in B_T(a(Z), K) \right]\right] \\
        &+ \ee_Z \ee_{g_t}\left[ \sum_{t = 1}^T \frac{d p_t}{d \mu}(Z) g_t(Z) \bbI\left[ t \not\in B_T(a(Z), K) \right] \right] \\
        &\leq \frac 1\sigma \ee_Z\left[ \sum_{t = 1}^T \bbI\left[ t \in B_T(a(Z), K) \right] \right] + \ee_Z\ee_{g_t}\left[ \sum_{t=  1}^T K \ptil_t(Z) g_t(Z) + \frac{K}{\sigma t}\right]  \\
        &\leq \frac{\epsilon T}{\sigma} + \frac{K \log(T)}{\sigma} + K \cdot \ee_{g_t}\left[ \sum_{t = 1}^T \frac 1t \cdot \sum_{s= 1}^{t-1} g_t(X_s') \right].\label{eq:decoupling_proof_balance}
    \end{align}
    The result follows by setting $\epsilon = \sqrt{\frac{2 }{T} \cdot \ee\left[ \sum_{t = 1}^T \frac 1t \sum_{s = 1}^{t-1} g_t(X_s') \right]}$.  \iftoggle{colt}{\jmlrQED}{\qed}

We remark that as mentioned earlier, in the case that $\sigma = 1$, the $a_t(Z) = 1$ uniformly over $Z$ and thus $B_T(a(Z), K) = \emptyset$ for all $K > 1$. In particular, this allows us to take $K$ a constant and $\epsilon \downarrow 0$ in \eqref{eq:decoupling_proof_balance} and recover the expected $\ee\left[ \sum_{t = 1}^T g_t(X_t) \right] \lesssim \ee\left[ \sum_{t = 1}^T g_t(X_t') \right]$ whenever the $X_t$ are iid

\subsection{Proof of Lemma \ref{lem:basic_inequality}}\label{app:wills_functional}
For the sake of simplicity, we drop the subscript from the notation for the ERM in this proof.  We begin by observing that because $\fstar \in \cF$, it holds by construction that
\begin{align}
    0 \leq \norm{\fstar - Y}_{T-1}^2 - \norm{\fhat - Y}_{T-1}^2.
\end{align}
Expanding the squares and rearranging then tells us that
\begin{align}
    0 \leq 2 \cdot \inprod{Y - \fstar}{\fhat - \fstar}_{T-1} - \norm{\fhat - \fstar}_{T-1}^2,
\end{align}
where $\inprod{\cdot}{\cdot}_{T-1}$ denotes the $L^2$ inner product with respect to the empirical measure on $X_1, \dots, X_{T-1}$.  Rearranging and observing that $Y - \fstar = \eta$ then tells us that
\begin{align}
    \frac 12 \cdot \norm{\fhat - \fstar}_{T-1}^2 \leq 2 \cdot \inprod{\eta}{\fhat - \fstar}_{T-1} - \frac 12 \cdot \norm{\fhat - \fstar}_{T-1}^2
\end{align}
and so
\begin{align}
    \norm{\fhat - \fstar}_{T-1}^2 \leq 4 \cdot \inprod{\eta}{\fhat - \fstar}_{T-1} -  \norm{\fhat - \fstar}_{T-1}^2
\end{align}
    Letting $\cG = \cF - \fstar$, we see that
    \begin{align}
        \ee\left[ \norm{\fhat - \fstar}_{T-1}^2 \right] &\leq  \ee\left[\sup_{g \in \cG} 4 \cdot \inprod{\eta}{g}_{T-1} - \norm{g}_{T-1}^2 \right] \\
        &\leq \ee\left[\sup_{g \in \cG} 4 \cdot \inprod{\eta - \eta'}{g}_{T-1} - \norm{g}_{T-1}^2 \right] \\
        &= \ee\left[\sup_{g \in \cG} 4 \cdot \inprod{\epsilon \cdot \abs{\eta - \eta'}}{g}_{T-1} - \norm{g}_{T-1}^2 \right] \\
        &\leq \ee\left[\sup_{g \in \cG} 8 \cdot \inprod{\epsilon \cdot \abs{\eta}}{g}_{T-1} - \norm{g}_{T-1}^2 \right] 
    \end{align}
    where $\epsilon$ is a vector of independent standard Rademacher random variables, and the second inequality follows from Jensens' and the fact that the $\eta$ are conditionally mean zero.  The final inequality above follows by the triangle inequality.  Now, by \Cref{lem:subgaussian_max}, we see that with probability at least $1 - \delta$, it holds that $\abs{\eta_t} \leq 2\nu \cdot \sqrt{\log\left( \frac T\delta \right)}$.  Observing that convex functions are extremized on the boundaries of convex sets, we see that
    \begin{align}
        \ee\left[\sup_{g \in \cG} 8 \cdot \inprod{\epsilon \cdot \abs{\eta}}{g}_{T-1} - \norm{g}_{T-1}^2 \right] &\leq 2 \nu \cdot \sqrt{\log\left( \frac T\delta \right)} \cdot \ee\left[\sup_{g \in \cG} 8 \cdot \inprod{\epsilon}{g}_{T-1} - \norm{g}_{T-1}^2 \right]  + 8 T \delta.
    \end{align}
    We now continue by controlling the expectation above.  Letting $\xi$ denote a vector of independent standard normal random variables, we see that 
    \begin{align}
        \ee\left[\sup_{g \in \cG} 8 \cdot \inprod{\epsilon }{g}_{T-1} - \norm{g}_{T-1}^2 \right] &\leq \ee\left[ \sup_{g \in \cG} 8\cdot \inprod{\xi }{g}_{T-1} - \norm{g}_{T-1}^2 \right],
    \end{align}
    again by Jensens' inequality and the fact that the sign and magnitude of a standard Gaussian are independent.  Now, let $\cE$ denote the high probability event from \Cref{lem:coupling} and observe that
    \begin{align}
        \ee\left[ \sup_{g \in \cG} 8\cdot \inprod{\xi}{g}_{T-1} - \norm{g}_{T-1}^2 \right] &= \ee\left[\bbI[\cE] \cdot  \sup_{g \in \cG} 8\cdot \inprod{\xi}{g}_{T-1} - \norm{g}_{T-1}^2 \right] + \ee\left[ \bbI[\cE^c] \cdot \sup_{g \in \cG} 8\cdot \inprod{\xi }{g}_{T-1} - \norm{g}_{T-1}^2 \right] \\
        &\leq \ee\left[\bbI[\cE] \cdot  \sup_{g \in \cG} 8\cdot \inprod{\xi }{g}_{T-1} - \norm{g}_{T-1}^2 \right] + 16 T \cdot e^{- \sigma k},
    \end{align}
    where the inequality follows from the bound on $\pp(\cE^c)$ from \Cref{lem:coupling} along with the independence of $\xi$ from $\cE$ and the fact that $\cF$ is uniformly bounded.  By Jensen's inequality, it holds for any $\lambda > 0$ that
    \begin{align}
        (T-1) \cdot &\ee\left[\bbI[\cE] \cdot  \sup_{g \in \cG} 8\cdot \inprod{\xi}{g}_{T-1} - \norm{g}_{t-1}^2 \right] \\
        &\leq \frac 1\lambda \cdot \log \ee\left[\bbI[\cE] \cdot  \exp\left(\lambda \cdot \sup_{g \in \cG} 8(T-1)\cdot \inprod{\xi }{ \cdot g}_{T-1} - (T-1) \cdot \norm{  g}_{T-1}^2  \right) \right].
    \end{align}
    Observing that $\lambda \geq \frac 1{32}$, we see that by \Cref{lem:monotonicity}, it holds that
    \begin{align}
        \frac 1\lambda &\cdot \log \ee\left[\bbI[\cE] \cdot  \exp\left(\lambda \cdot \sup_{g \in \cG} 8(T-1)\cdot \inprod{\xi }{ \cdot g}_{T-1} - (T-1) \cdot \norm{  g}_{T-1}^2  \right) \right] \\
        &\leq \frac 1\lambda \cdot \log \ee\left[\bbI[\cE] \cdot  \exp\left(\lambda \cdot \sup_{g \in \cG} 8\cdot \sum_{s = 1}^{T-1}\sum_{j  =1}^k \xi_{s,j} \cdot g(Z_{s,j}) -  g(Z_{s,j})^2  \right) \right] \\
        &\leq \frac 1\lambda \cdot \log \ee\left[ \cdot  \exp\left(\lambda \cdot \sup_{g \in \cG} 8\cdot \sum_{s = 1}^{T-1}\sum_{j  =1}^k \xi_{s,j} \cdot g(Z_{s,j}) -  g(Z_{s,j})^2  \right) \right].
    \end{align}
    Setting $\lambda = \frac 1{32}$, now, and dividing by $T-1$, we see that
    \begin{align}
        \ee\left[\bbI[\cE] \cdot \sup_{g \in \cG} 8 \cdot \inprod{\epsilon }{g}_{T-1} - \norm{g}_{T-1}^2 \right] &\leq \frac{32}{T-1} \cdot \log \ee_{Z_{s,j}}\left[ W_{k(T-1)}\left( 256 \cdot \cG \right) \right].
    \end{align}
    The result follows immediately. \iftoggle{colt}{\jmlrQED}{\qed}

\subsection{Concluding the Proof}

Applying \Cref{lem:decoupling} with $g_t = (\fhat_t - \fstar)^2$, we see that
    \begin{align}
        \ee\left[ \sum_{t = 1}^T \left( f_t(X_t) - \fstar(X_t) \right)^2  \right] &\leq \frac{\log^2(T)}{\sigma} \cdot \sqrt{2T \cdot \ee\left[ \sum_{t = 1}^T \frac{1}{t} \cdot \sum_{s = 1}^{t-1} \left( \fhat_t(X_s') - \fstar(X_s') \right)^2 \right]} \\
        &= \frac{\log^2(T)}{\sigma} \cdot \sqrt{2 T \cdot \sum_{t = 1}^T \frac 1t \cdot \ee\left[  \cdot \sum_{s = 1}^{t-1} \left( \fhat_t(X_s') - \fstar(X_s') \right)^2 \right]}.\label{eq:concluding_proof1}
    \end{align}
    We now compute for each $t \in [T]$,
    \begin{align}
        \frac 1t \cdot \ee\left[ \sum_{s = 1}^{t-1} \left( \fhat_t(X_s') - \fstar(X_s') \right)^2 \right] &\leq 2 \cdot \ee\left[ \sum_{s = 1}^{t-1} \left( \fhat_t(X_s) - \fstar(X_s) \right)^2 \right] \\
        &\quad+ \frac 1t \cdot\ee\left[ \sup_{f \in \cF} \sum_{s = 1}^{t-1} \left( f(X_s') - \fstar(X_s') \right)^2  - 2 \cdot \sum_{s = 1}^{t-1}\left( f(X_s) - \fstar(X_s) \right)^2\right] \\
        &\leq 2 \cdot \ee\left[ \norm{\fhat_t - \fstar}_{t-1}^2 \right] \\
        &\quad+ \frac 1t \cdot\sqrt{\frac \pi 2} \cdot \frac 92 \cdot \log \ee_\mu\left[ W_{2(t-1)\log(t-1)/\sigma}\left( 4 \cdot (\cF - \fstar) \right) \right] + \frac 6t \\
        &\leq 2 \cdot \ee\left[ \norm{\fhat_t - \fstar}_{t-1}^2 \right] + \frac 9t \cdot \log \ee_\mu \left[ W_{2T\log(T)/\sigma}(4 \cdot \cF ) \right] + \frac 6t,
    \end{align}
    where the first inequality follows because $\fhat_t \in \cF$, the second inequality is \Cref{thm:tight_norm_comparison}, and the final inequality follows because $W_m(\cF)$ is monotone in $m$ and invariant under translation.  By \Cref{lem:basic_inequality}, we have that 
    \begin{align}
        \ee\left[ \norm{\fhat_t - \fstar}_{t-1}^2 \right] \leq \frac{64}{t} \cdot \nu \cdot \sqrt{\log(T)} \left( \log \ee_\mu\left[ W_{2T\log(T)/\sigma}\left( 256 \cdot \cF \right) \right] + \frac 1t \right).
    \end{align}
Combining this with the previous display implies that
\begin{align}
    \frac 1t \cdot \ee\left[ \sum_{s = 1}^{t-1} \left( \fhat_t(X_s') - \fstar(X_s') \right)^2 \right] &\leq \frac{150 (1+\nu)}{t} \cdot \sqrt{\log(T)} \left( 1 + \log \ee_\mu\left[ W_{2T\log(T)/\sigma}(256 \cdot \cF) \right] \right)
\end{align}
and thus
\begin{align}
    \sum_{t = 1}^T \frac 1t \cdot \ee\left[ \sum_{s = 1}^{t-1} \left( \fhat_t(X_s') - \fstar(X_s') \right)^2 \right]  &\leq 150 (1+\nu) \log^{3/2}(T)\left( 1 +  \log \ee_\mu\left[ W_{2T\log(T)/\sigma}(256 \cdot \cF) \right]\right).
\end{align}
Plugging this into \eqref{eq:concluding_proof1} concludes the proof. \iftoggle{colt}{\jmlrQED}{\qed}
\section{Proof of Theorem \ref{thm:tight_norm_comparison}}\label{app:sharp_norm_comparison}
This appendix is devoted ot the proof of the sharp norm comparison result, \Cref{thm:tight_norm_comparison}.  We begin by rearranging the sum to observe that
    \begin{align}
        \ee\left[ \sup_{f \in \cF} \sum_{t = 1}^T f(X_t') - (1+2c) \cdot f(X_t)^2  \right] &= \ee\left[ \sup_{f \in \cF} \sum_{t = 1}^T (1+c) (f(X_t')^2 -  f(X_t)^2) - c f(X_t')^2 - c f(X_t)^2 \right].\label{eq:norm_comparison_pf_1}
    \end{align}
    We now take inspiration from \citet{rakhlin2011online} and consider the following \emph{tree of probabilities} construction.  For $t \in [T]$, let $p_t(\cdot | x_1, \dots, x_{t-1})$ be the distribution of $X_t$ conditioned on the history that $X_s = x_s$ for $s < t$.  For $x, x' \in \cX$ and $\epsilon \in \left\{ \pm1 \right\}$, define the selector function
    \begin{align}
        \chi(x,x', \epsilon) = \begin{cases}
            x & \epsilon = -1 \\
            x' & \epsilon = 1
        \end{cases}\label{eq:selector}
    \end{align}
    and write $\chi_t(\epsilon)$ for the $t$-th selector when $x_t, x_t'$ are clear from context.  We form the following tree of probabilities $\rho$, where we associate for each path $\epsilon \in \left\{ \pm 1 \right\}^T$, the measure $\rho_t(\epsilon_{1:t-1})$ on pairs $(X_t, X_t')$ conditional on $(X_{1:t-1}, X_{1:t-1}')$ such that
    \begin{align}
        \rho_t(\epsilon_{1:t-1})((X_1,X_1'), \dots, (X_{t-1}, X_{t-1}')) = (p_t(\cdot | \chi_1(\epsilon_1), \dots, \chi_{t-1}(\epsilon_{t-1})), p_t(\cdot | \chi_1(\epsilon_1), \dots, \chi_{t-1}(\epsilon_{t-1}))).\label{eq:tree_of_probs}
    \end{align}
    In words, $\rho_t$ on a fixed path $\epsilon$ is a conditional measure that samples $(X_t, X_t')$ inependently from $p_t$ conditioned on an $\epsilon$-dependent history.  In this way, if we let $\rho = (\rho_1, \dots, \rho_T)$, we have a measure on two coupled $\cX$-valued complete binary trees of depth $T$.  For more exposition on such trees of probabilities, we refer the reader to \citet[\S 3]{rakhlin2011online}.  
    
    Continuing in the proof, we now write for simplicity for all $1 \leq s \leq s' \leq T$,
    \begin{align}
        S_{s:s'}(f) = \sum_{t = s}^{s'} f(X_t)^2 \qquad \text{and} \qquad S_{s:s'}'(f) = \sum_{t = s}^{s'} f(X_t')^2.
    \end{align}
    Writing out the expectations in the right hand side of \eqref{eq:norm_comparison_pf_1}, we observe that it is equal to
    \begin{align}
        \ee_{X_1, X_1' \sim p_1} \ee_{X_2, X_2' \sim p_2(\cdot | X_1)} \cdots \ee_{X_T, X_T' \sim p_T(\cdot | X_{1:T-1})}\left[ \sup_{f \in \cF} (1+c)(S_{1:T}'(f) - S_{1:T}(f)) - c(S_{1:T}'(f) + S_{1:T}(f)) \right].
    \end{align}
    We now observe that if we switch the role of $X_1$ and $X_1'$, then we have by symmetry that the above expectation is equal to
    \begin{align}\label{eq:norm_comparison_pf_2}
       &\ee_{X_1', X_1 \sim p_1} \ee_{X_2, X_2' \sim p_2(\cdot | X_1')} \cdots \\
        &\cdots\ee_{X_T, X_T' \sim p_T(\cdot | X_1', X_{2:T})}\left[ \sup_{f \in \cF} (1+c)(-(f^2(X_1') - f^2(X_1)) +  S_{2:T}'(f) - S_{2:T}(f)) - c(S_{1:T}'(f) + S_{1:T}(f)) \right],
    \end{align}
    where we emphasize that the subtracted term is \emph{symmetric} with respect to exchanging $X_t$ for $X_t'$ as opposed to antisymmetric.  In particular, if we define
    \begin{align}
        \chibar(x,x',\epsilon) = \begin{cases}
            x' & \epsilon = -1 \\
            x & \epsilon = 1
        \end{cases},
    \end{align}
    the opposite of the $\chi$ in \eqref{eq:selector} and we use a similar abbreviation $\chibar_t(\epsilon)$ for the $t$-th selector, then we may continue in the same way as \eqref{eq:norm_comparison_pf_2} and observe that for any $\epsilon_{1:T} \in \left\{ \pm 1 \right\}^T$, that the expectation in the right hand side of \eqref{eq:norm_comparison_pf_1} is equal to
    \begin{align}
        &\ee_{X_1,X_1' \sim p_1} \ee_{X_2, X_2' \sim p_2(\cdot | \chi_1(\epsilon_1))}\cdots\\
        &\cdots \ee_{X_T, X_T' \sim p_T(\cdot | \chi_1(\epsilon_1), \dots, \chi_{T-1}(\epsilon_{T-1}))}\left[ \sup_{f \in \cF} \sum_{t =1}^T \epsilon_t (1+c) \left( f^2(\chi_t(\epsilon)) - f^2(\chibar_t(\epsilon)) \right) - c \left( f^2(\chibar_t(\epsilon)) + f^2(\chi_t(\epsilon)) \right)\right].
    \end{align}
    Because this equality holds true for all choices of signs $\epsilon$, we may take an expectation over the distribution that is uniform on the signs and observe that the preceding display is equal to
    \begin{align}
        &\ee_{X_1, X_1'\sim p_1} \ee_{\epsilon_1} \ee_{X_2, X_2' \sim p_2(\cdot | \chi_1(\epsilon_1))} \ee_{\epsilon_2} \cdots \\
        &\cdots \ee_{X_T, X_T' \sim p_T(\cdot | \chi_1(\epsilon_1), \dots \chi_{T-1}(\epsilon_{T-1}))}\ee_{\epsilon_T}\left[ \sup_{f \in \cF} \sum_{t =1}^T \epsilon_t (1+c) \left( f^2(\chi_t(\epsilon)) - f^2(\chibar_t(\epsilon)) \right) - c \left( f^2(\chibar_t(\epsilon)) + f^2(\chi_t(\epsilon)) \right)\right] \\
        &= \ee_{(\bx, \bx') \sim \rho}\ee_\epsilon\left[ \sup_{f\in \cF} \sum_{t= 1}^T (1+c)\epsilon_t\left( f^2(\bx_t(\epsilon)) - f^2(\bx_t'(\epsilon)) \right) - c \left( f^2(\bx_t(\epsilon)) + f^2(\bx_t'(\epsilon)) \right) \right],
    \end{align}
    where the $\rho$ is from \eqref{eq:tree_of_probs}, which forms a measure on coupled $\cX$-valued complete binary trees of depth $T$.  Now we may split the supremum in two and use the symmetry of the Rademacher distribution to conclude that
    \begin{align}
        \ee_{(\bx, \bx') \sim \rho}&\ee_\epsilon\left[ \sup_{f\in \cF} \sum_{t= 1}^T (1+c)\epsilon_t \left( f^2(\bx_t(\epsilon)) - f^2(\bx_t'(\epsilon)) \right) - c \left( f^2(\bx_t(\epsilon)) + f^2(\bx_t'(\epsilon)) \right) \right] \\
        &\leq \ee_{(\bx, \bx') \sim \rho}\ee_\epsilon\left[ \sup_{f\in \cF} \sum_{t= 1}^T (1+c)\epsilon_t  f^2(\bx_t(\epsilon))  - c  f^2(\bx_t(\epsilon))  \right] \\
        &\quad + \ee_{(\bx, \bx') \sim \rho}\ee_\epsilon\left[ \sup_{f\in \cF} \sum_{t= 1}^T -(1+c)\epsilon_t  f^2(\bx_t'(\epsilon))  - c  f^2(\bx_t'(\epsilon))  \right] \\
        &\leq 2 \cdot \ee_{(\bx, \bx') \sim \rho}\ee_\epsilon\left[ \sup_{f\in \cF} \sum_{t= 1}^T (1+c)\epsilon_t  f^2(\bx_t(\epsilon))  - c  f^2(\bx_t(\epsilon))  \right]. \label{eq:norm_comparison_pf_3}
    \end{align}
    Above, the first inequality follows by Jensens' and the second follows by symmetry.  More precisely, for the second inequality, we observe that
    \begin{align}
        \ee\left[ \sup_{f\in \cF} \sum_{t= 1}^T -(1+c)\epsilon_t  f^2(\bx_t'(\epsilon))  - c  f^2(\bx_t'(\epsilon))  \right] &= \ee\left[ \sup_{f\in \cF} \sum_{t= 1}^T (1+c)\epsilon_t  f^2(\bx_t'(-\epsilon))  - c  f^2(\bx_t'(-\epsilon))  \right] \\
        &= \ee\left[ \sup_{f\in \cF} \sum_{t= 1}^T (1+c)\epsilon_t  f^2(\bx_t(\epsilon))  - c  f^2(\bx_t(\epsilon))  \right],
    \end{align}
    with the second equality following because $\chibar(-\epsilon_t) = \chi(\epsilon_t)$, the Rademacher distribution is symmetric, and $\bx_t'(\epsilon), \bx_t(\epsilon)$ are identically distributed.

    We now proceed to bound the right hand side of \eqref{eq:norm_comparison_pf_3}.
    Noting that $\bx_t(\epsilon)$ is $\sigma$-smooth with respect to $\mu$ conditioned on the history for all $t \in [T]$, we may apply \Cref{lem:tree_coupling} and observe that for fixed $k$, there is some event $\cE$ under which we may sample $Z_{t,j}, Z_{t,j}' \sim \mu$ independent for $1 \leq j \leq k$ and it holds that $\bx_t(\epsilon) \in \left\{ Z_{t,j} | j \in [k] \right\}$ for all $t \in [T]$ and similarly for $Z_{t,j'}$ and $\bx_t'(\epsilon)$; furthermore $\pp\left( \cE^c \right) \leq 2T e^{-\sigma k}$.  Thus we observe that under this coupling $\Pi$,
    \begin{align}
        2 \cdot &\ee_{(\bx, \bx') \sim \rho}\ee_\epsilon\left[ \sup_{f\in \cF} \sum_{t= 1}^T (1+c)\epsilon_t  f^2(\bx_t(\epsilon))  - c  f^2(\bx_t(\epsilon))  \right] \\
        &= 2 \cdot \ee_{(\bx, \bx'), Z_{t,j}, \epsilon \sim \Pi}\left[ \bbI[\cE] \cdot \sup_{f\in \cF} \sum_{t= 1}^T (1+c)\epsilon_t  f^2(\bx_t(\epsilon))  - c  f^2(\bx_t(\epsilon))  \right] \\
        &\quad +2 \cdot \ee_{(\bx, \bx'), Z_{t,j}, \epsilon \sim \Pi}\left[\bbI[\cE^c]\cdot \sup_{f\in \cF} \sum_{t= 1}^T (1+c)\epsilon_t  f^2(\bx_t(\epsilon))  - c  f^2(\bx_t(\epsilon))  \right] \\
        &\leq 2 \cdot\ee_{(\bx, \bx'), Z_{t,j}, \epsilon \sim \Pi}\left[ \bbI[\cE] \cdot \sup_{f\in \cF} \sum_{t= 1}^T (1+c)\epsilon_t  f^2(\bx_t(\epsilon))  - c  f^2(\bx_t(\epsilon))  \right] + 4(1+c)T^2 e^{- \sigma k}.
    \end{align}
    Letting $\xi_t$ be a standard Gaussian, we may apply Jensen's inequality to conclude that
    \begin{align}
        \ee_{(\bx, \bx'), Z_{t,j}, \epsilon \sim \Pi}&\left[ \bbI[\cE] \cdot \sup_{f\in \cF} \sum_{t= 1}^T (1+c)\epsilon_t  f^2(\bx_t(\epsilon))  - c  f^2(\bx_t(\epsilon))  \right] \\
        &= \ee_{(\bx, \bx'), Z_{t,j}, \epsilon \sim \Pi}\left[ \bbI[\cE] \cdot \sup_{f\in \cF} \sum_{t= 1}^T (1+c)\epsilon_t \frac{\ee\left[ \abs{\xi_t} \right]}{\ee\left[ \abs{\xi_t} \right]}  f^2(\bx_t(\epsilon))  - c  f^2(\bx_t(\epsilon))  \right] \\
        &\leq \sqrt{\frac \pi 2} \cdot \ee_{(\bx, \bx'), Z_{t,j}, \xi \sim \Pi'}\left[ \bbI[\cE] \cdot \sup_{f\in \cF} \sum_{t= 1}^T (1+c)\xi_t  f^2(\bx_t(\xi))  - c  f^2(\bx_t(\xi))  \right],
    \end{align}
    where $\Pi'$ is the coupling $\Pi$, but replacing $\epsilon_t$ with $\xi_t = \epsilon_t \cdot \abs{\xi_t'}$ for $\xi_t'$ independent standard Gaussians.
    Above, we we used the fact that a Gaussian's norm and sign are independent and we abused notation by letting $\bx_t(\xi) = \bx_t(\sign(\xi))$.  Combining the results thus far and observing that $f^4(x) \leq f^2(x)$ for all $x \in \cX$, we have shown that for any $k \in \bbN$ and $c > 0$,
    \begin{align}
        \ee_{(\bx, \bx'), Z_{t,j}, \xi \sim \Pi'}&\left[ \sup_{f \in \cF} \sum_{t = 1}^T f(X_t') - (1+2c) \cdot f(X_t)^2  \right] \\
        &\leq \sqrt{2 \pi} \cdot(1 + c) \cdot \ee_{(\bx, \bx'), Z_{t,j}, \xi \sim \Pi'}\left[ \bbI[\cE] \cdot \sup_{f\in \cF} \sum_{t= 1}^T \xi_t  f^2(\bx_t(\xi))  - \frac{c}{1+c}  f^4(\bx_t(\xi))  \right] + 4 T^2 \cdot e^{-\sigma k}.
    \end{align}
    To conclude the proof, we apply Jensen's inequality and observe that for any $\lambda > 0$, it holds that
    \begin{align}
        \ee_{(\bx, \bx'), Z_{t,j}, \xi \sim \Pi'}&\left[ \bbI[\cE] \cdot \sup_{f\in \cF} \sum_{t= 1}^T \xi_t  f^2(\bx_t(\xi))  - \frac{c}{1+c}  f^4(\bx_t(\xi))  \right] \\
        &\leq \frac 1\lambda \cdot \log  \ee_{(\bx, \bx'), Z_{t,j}, \xi \sim \Pi'}\left[ \exp\left( \bbI[\cE] \cdot \sup_{f\in \cF} \sum_{t= 1}^T \xi_t \lambda  f^2(\bx_t(\xi))  - \frac{c\lambda }{1+c}  f^4(\bx_t(\xi))  \right)\right].
    \end{align}
    Setting $\lambda = \frac{2c}{1 + c}$ and applying \Cref{lem:monotonicity} then implies that
    \begin{align}
        \frac 1\lambda &\cdot\log  \ee_{(\bx, \bx'), Z_{t,j}, \xi \sim \Pi'}\left[ \exp\left( \bbI[\cE] \cdot \sup_{f\in \cF} \sum_{t= 1}^T \xi_t \lambda  f^2(\bx_t(\xi))  - \frac{c\lambda }{1+c}  f^4(\bx_t(\xi))  \right)\right] \\
        &\leq \frac{1+c}{2c} \cdot \log \ee_{(\bx, \bx'), Z_{t,j}, \xi \sim \Pi'}\left[ \exp\left( \sup_{f \in \cF} \sum_{t = 1}^T \sum_{j =1}^k \xi_{t,j} \left( \sqrt{\frac{2c}{1 + c}}\cdot f(Z_{t,j}) \right)^2 - \frac 12 \cdot \left( \sqrt{\frac{2c}{1+c}} f(Z_{t,j}) \right)^4  \right) \right] \\
        &= \frac{1 + c}{2 c} \cdot \log  \ee_{Z_{t,j} \sim \mu} \ee_\xi \left[ \exp\left( \sup_{f \in \cF} \sum_{t = 1}^T \sum_{j =1}^k \xi_{t,j} \left( \sqrt{\frac{2c}{1 + c}}\cdot f(Z_{t,j}) \right)^2 - \frac 12 \cdot \left( \sqrt{\frac{2c}{1+c}} f(Z_{t,j}) \right)^4  \right) \right] \\
        &= \frac{1+c}{2c} \cdot \log  \ee_{Z_{t,j} \sim \mu} W_{kT}\left( \frac{2c}{1+c} \cdot \cF^2 \right),
    \end{align}
    where $W_{kT}$ is the Will's functional defined in \Cref{def:wills}.  We now note that because $\cF$ is uniformly bounded, it holds that $f \mapsto f^2$ is 2-Lipschitz and we may apply \citet[Theorem 4.1]{mourtada2023universal} to yield that $ W_{kT}\left( \frac{2c}{1+c} \cdot \cF^2 \right) \leq  W_{kT}\left( \frac{4c}{1+c} \cdot \cF \right)$.  Putting everything together yields
    \begin{align}
        \ee\left[ \sup_{f \in \cF} \sum_{t = 1}^T f^2(X_t') - (1+2c) \cdot f^2(X_t)  \right] \leq \sqrt{\frac \pi 2} \cdot \frac{(1+c)^2}{c} \cdot \log  \ee_{Z_{t,j} \sim \mu} W_{kT}\left( \frac{2c}{1+c} \cdot \cF^2 \right) + 4(1+c)T^2 e^{-\sigma k}.
    \end{align}
    Setting $k = 2 \log(T) / \sigma$ concludes the proof. \iftoggle{colt}{\jmlrQED}{\qed}

Finally, we state the form of the coupling result (\Cref{lem:coupling}) that we require in the above proof.
\begin{lemma}[Lemma 24 from \citet{block2022smoothed}]\label{lem:tree_coupling}
    Let $p_t(\cdot | x_{1:t-1})$ denote the conditional distribution of $X_t$ given the history and let $\rho$ be the measure on the pair $(\bx, \bx')$ of $\cX$-labelled-complete binary trees defined in \eqref{eq:tree_of_probs}.  If $p_t$ is $\sigma$-smooth with respect to $\mu$ for all $t \in [T]$, then for all $k \in \bbN$, there exists a coupling $\Pi$ among $\epsilon_{1:T}$, $(\bx, \bx')$, and $\left\{ Z_{t,j}, Z_{t,j}' | t \in [T], \, j \in [k] \right\}$ such that the following properties hold:
    \begin{enumerate}
        \item The $\epsilon_{1:T}$ are independent Rademacher random variables.
        \item The $Z_{t,j}, Z_{t,j}' \sim \mu$ are independent samples from $\mu$.
        \item The $(\bx, \bx') \sim \rho$.
        \item $\epsilon_{1:T}$ is independent of $\left\{ Z_{t,j}, Z_{t,j'} \right\}$.
        \item There is an event $\cE$ with probability at least $1 - 2 T e^{- \sigma k}$ such that on $\cE$, $\bx_t(\epsilon) \in \left\{ Z_{t,j} | j \in [k] \right\}$ and $\bx_t'(\epsilon) \in \left\{ Z_{t,j} | j \in [k] \right\}$ for all $t \in [T]$.
    \end{enumerate}
\end{lemma}
\section{Proof of Theorem \ref{thm:erm_lowerbound}}\label{app:erm_lowerbound}

In this appendix, we prove the lower bound of \Cref{thm:erm_lowerbound}.  Fix $d \in \bbN$ and let $\cX = [0,1]^d \subset \rr^d$.  We let
\begin{align}
    \cF = \left\{ x \mapsto \min\left( \bbI[x_i \geq \theta_i] \right) | \theta_i \in [0,1]^d \right\}
\end{align}
be the class of $d$-dimensional axis-aligned thresholds.  It is classical that $\vc(\cF) = 2d$ \citep{van2014probability,mohri2018foundations} and thus $\rad_m(\cF) \lesssim \sqrt{d m}$ for all $m$.  Let $\fstar = 0$ and define the $\erm$ as follows.  Given a data set of $(X_1, Y_1), \dots, (X_t, Y_t)$, let
\begin{align}
    \cF_{(X_{1:t},Y_{1:t})} = \left\{ f \in \cF | \norm{f(X)- Y}_t = \min_{f' \in \cF} \norm{f'(X) - Y}_t \right\}
\end{align}
be the set of minimizers of the empirical risk.  Note that this set is always nonempty due to the compactness of $\cF$ and the continuity of the norm.  For each $t$, and each coordinate $i$, we will let $\theta_{t,i} = \inf_{\theta \in \cF_{(X_{1:t-1}, Y_{1:t-1})}} \theta_i$ denote the minimal threshold in the $i$-th coordinate that still minimizes the empirical risk.  We let the data be realizable and thus $Y_t = 0$ for all $t \in [T]$.  We claim that for any $\epsilon > 0$ there exists an adversary forcing the above defined ERM to get
\begin{align}
    \ee\left[ \reg_T \right] \geq \frac 12 \cdot \min\left( \frac{1-\sigma^{1/d}}{\epsilon} \cdot d, \frac{\epsilon T}{\sigma^{1/d}} \right).
\end{align}
We construct the adversary as follows.  We introduce the sequence of stopping times $\tau_{i, j}$ for $i \in [d]$ and $j \in \bbN$ as follows.  Let $\tau_{1,0} = 0$ and for $i,j > 0$, let 
\begin{align}
    \tau_{i,j} = \inf\left\{ t > 0 | \max_{s \leq t} X_{s,i} \geq 1 - \sigma^{1/d} + (j-1) \epsilon \right\}.
\end{align}
For $i > 1$, let $\tau_{i,0} = \tau_{i-1, \lfloor (1-\sigma^{1/d}) / \epsilon \rfloor}$.  In words, $\tau_{i,j}$ is the first time that the $i$-th coordinate of the data exceeds $1 - \sigma^{1/d} + (j-1) \epsilon$ and $\tau_{i,0}$ is the first time that the $(i-1)^{st}$ coordinate has exceeded $1 - \epsilon$.  For any $t$, let $\tau(t) = \tau_{i_t,j_t}$, where $i_t = \argmax_{i \in [d]} \tau_{i,0} \leq t$ and $j_t = \argmax_{j \in \bbN} \tau_{i_t, j} \leq t$.  

We now define the distributions of the $X_t$.  Let $p_{j}$ be a distribution on $[0,1]$ such that $p_j = \unif\left( \left[  j \epsilon, \sigma^{1/d} +j \epsilon \right] \right)$ for $j \leq \frac{1 - \sigma^{1/d}}{\epsilon}$.  Finally, we let
\begin{align}
    P_t = \left( \bigotimes_{i = 1}^{i_{t-1} - 1} p_0  \right) \otimes p_{j_{t-1}} \otimes \left( \bigotimes_{i = i_{t-1} + 1}^{d} p_0 \right).
\end{align}
In words, if $X_t \sim P_t$, then the coordinates of $X_t$ are independent and distributed uniformly in $[0, \sigma^{1/d}]$ except for the $i_{t-1}$-th coordinate, which is distributed uniformly in $[j_{t-1} \epsilon, j_{t-1} \epsilon + \sigma^{1/d}]$.  We reiterate that $Y_t = 0$ uniformly.

We observe that $P_t$ is $\sigma$-smooth with respect to $\unif\left( [0,1]^d \right)$; indeed, for any $t$, it holds that $P_t$ is uniform on a body of volume $\sigma$ contained in $[0,1]^d$.  Thus it suffices to show that the expected number of times that $\fhat_t(X_t) = 1$ is large.  Observe that by construction of the ERM, it holds that $\fhat_t(X_t) = 1$ if and only if at least one coordinate of $X_t$ is strictly larger than the previous largest observed data point in that coordinate, i.e., if there exists some $i \in [d]$ such that $X_{t,i} > \max_{s < t} X_{s,i}$.  By construction of $P_t$, then, it holds that
\begin{align}
    \ee\left[ \reg_T \right] &= \ee\left[ \sum_{t = 1}^T \max_{i \in [d]} \bbI\left[X_{t,i} > \max_{s < t} X_{s,i}  \right] \right] \geq \ee\left[ \sum_{t=  1}^T \bbI[\tau(t) \neq \tau(t-1)] \right].
\end{align}
We now observe that as long as $\tau(t-1) < \tau_{d, \lfloor (1-\sigma^{1/d}) / \epsilon \rfloor}$, it holds by construction that
\begin{align}
    \pp(\tau(t) \neq \tau(t-1) | \tau(t-1)) = \begin{cases}
        \frac{\epsilon}{\sigma^{1/d}} & i_{t-1} \leq d \text{ or } j_{t-1} < \frac{1 - \sigma^{1/d}}{\epsilon} \\
        0 & \text{otherwise}.
    \end{cases}
\end{align}
Thus by by the tower law of conditional expectation, it holds that
\begin{align}
    \ee\left[ \sum_{t=  1}^T \bbI[\tau(t) \neq \tau(t-1)] \right] &= \frac{\epsilon T}{\sigma^{1/d}} \cdot \pp\left( \tau(T) < \tau_{d, \lfloor (1-\sigma^{1/d}) / \epsilon \rfloor} \right) + d \cdot \left\lfloor \frac{1 - \sigma^{1/d}}{\epsilon}\right\rfloor \cdot  \pp\left( \tau(T) = \tau_{d, \lfloor (1-\sigma^{1/d}) / \epsilon \rfloor} \right) \\
    &\geq \frac 12 \min\left( \frac{1-\sigma^{1/d}}{\epsilon} \cdot d, \frac{\epsilon T}{\sigma^{1/d}} \right).
\end{align}
Taking a maximum over $\epsilon$ concludes the proof.

\section{Miscellaneous Lemmata}\label{app:lemmata}

\begin{lemma}\label{lem:uniform_deviations}
    Let $\cF: \cX \to [-1,1]$ be a function class and  $X_t$ a sequence of $\sigma$-smoothed data with respect to $\mu$.  Then for any $k \in \bbN$, it holds that
    \begin{align}
        \ee\left[ \sup_{f \in \cF} \sum_{t = 1}^T f(X_t) - f(X_t') \right] \leq 2\rad_{k T}(\cF) + 2T^2 e^{- \sigma k}.
    \end{align}
\end{lemma}
\begin{proof}
    By \citet[Theorem 3]{rakhlin2011online}, it holds that
    \begin{align}
        \ee\left[ \sup_{f \in \cF} \sum_{t = 1}^T f(X_t) - f(X_t') \right] \leq 2 \cdot \sup_{\rho} \ee_\rho\left[ \sup_{f \in \cF} \sum_{t = 1}^T \epsilon_t f(X_t(\epsilon)) \right],
    \end{align}
    where $X_t(\epsilon)$ is a path of a $\cX$-valued binary tree distributed according to $\rho$, as defined in \citet{rakhlin2011online}.  By \citet[Lemma 17]{block2022smoothed}, however, it holds that for any $k \in \bbN$,
    \begin{align}
        \sup_{\rho} \ee_\rho\left[ \sup_{f \in \cF} \sum_{t = 1}^T \epsilon_t f(X_t(\epsilon)) \right] \leq \rad_{k T}(\cF) + T^2 e^{- \sigma k}.
    \end{align}
    The result follows immediately.
\end{proof}

\begin{lemma}[Lemma 5.2 in \citet{van2014probability}]\label{lem:subgaussian_max}
    Let $\eta_1, \dots, \eta_T$ denote a collection of possibly dependent random variables such that all $\eta_t$ are $\nu^2$-subGaussian.  Then for any $\delta > 0$, it holds with probability at least $1 - \delta$ that
    \begin{align}
        \max_{t \in [T]} \abs{\eta_t} \leq \nu \cdot \sqrt{2 \log\left( \frac{2T}{\delta} \right)}.
    \end{align}
\end{lemma}
\begin{lemma}\label{lem:monotonicity}
    Suppose that $\Psi, \psi: \cG \to \rr$ are two functionals and $B, \lambda > 0$ are two constants such that $\lambda  \geq \frac{2}{B^2}$.  Let $\cE$ be an event independent of $\xi \sim \cN(0,1)$.  Then it holds that
    \begin{align}
        \ee\left[\bbI[\cE] \cdot  \exp\left( \sup_{g \in \cG} \Psi(g) \right) \right] \leq \ee\left[\bbI[\cE] \cdot \exp\left( \sup_{g \in \cG} \Psi(g) + B \lambda \xi \psi(g) - \lambda \psi^2(g) \right)  \right]
    \end{align}
\end{lemma}
\begin{proof}
    Note that
    \begin{align}
        \ee_\xi\left[ e^{B \lambda \xi \psi(g) - \lambda \psi^2(g)} |\cE \right] = e^{\left( \frac{B^2 \lambda^2}{2} - \lambda \right) \psi^2(g)} \geq 1,
    \end{align}
    where the equality follows from the independence of $\cE$ and $\xi$ as well as the Gaussianity of the latter and the inequality follows from the assumption on $\lambda$.  The result follows immediately.
\end{proof}

\section{Stronger Norm Comparison Using the Small Ball Method}\label{app:small_ball}

We showed in \Cref{thm:tight_norm_comparison} that whenever a function class $\cF$ is bounded and the data $X_1,\dots, X_T$ are smooth, a sharp norm comparison holds, i.e.,
\begin{align}\label{eq:norm_comparison_example}
    \ee\left[ \sup_{f \in \cF} \norm{f}_{\ptil_T}^2 - (1 + c)\norm{f}_T^2  \right] \lesssim \frac{\comp(\cF) \log\left( \frac T\sigma \right)}{T},
\end{align}
where $\ptil_T = \frac 1T \sum_{t = 1}^T p_t$ and $p_t$ is the law of $X_t$.  In this appendix, we show that under a certain anti-concentration condition, a stronger norm comparison holds.  In particular, by definition of smoothness, if $p_t$ is smooth with respect to $\mu$ then for all functions $f$, it holds that $\norm{f}_{T}^2 \lesssim \norm{f}_\mu^2$.  In general, the reverse inequality does not hold, however, as witnessed by $p_t$ having support on some strict subset of $\cX$ and $f$ being the indicator of the complement.  We show that under a `small-ball' type condition, the reverse inequality does hold and, in fact, the norm $\norm{\cdot}_{\ptil_T}^2$ in \eqref{eq:norm_comparison_example} can be replaced by $\norm{\cdot}_\nu^2$ \emph{for any smooth measure } $\nu$.  This result amounts to a smoothed-data analogue of the celebrated small-ball argument of \citet{koltchinskii2015bounding,mendelson2015learning}.  We begin by stating the main result of this section.
\begin{theorem}\label{thm:small_ball}
    Suppose that $\cF: \cX \to \rr$ is a function class, $\mu \in \Delta(\cX)$ and $X_t \sim p_t$ are $\sigma$-smooth with respect to $\mu$ for $t \in [T]$.  Suppose further that there are constants $1 > c, c' > 0$ such that
    \begin{align}\label{eq:smallball}
        \sup_{f \in \cF} \mu\left( \abs{f(Z)} < \sqrt{\frac {2c}\sigma} \cdot \norm{f}_\mu \right) \leq \sigma (1 - c').
    \end{align}
    Let $\cN(\cF, \epsilon)$ denote the covering number of $\cF$ with respect to $\norm{\cdot}_\mu$ and suppose that for some constant $C > 0$, \iftoggle{colt}{$T \geq \frac{C}{\sigma} \cdot \log\abs{\cN\left( \cF, \frac{\sigma^2  \deltatil^2c c'}{C} \right)} \cdot  \log^3\left( \frac {C}{\sigma\deltatil cc'} \right)$.  }{
    \begin{align}
        T \geq \frac{C}{\sigma} \cdot \log\abs{\cN\left( \cF, \frac{\sigma^2  \deltatil^2c c'}{C} \right)} \cdot  \log^3\left( \frac {C}{\sigma\deltatil cc'} \right).
    \end{align}
    }
    Then for any measure $\nu$ that is $\sigma$-smooth with respect to $\mu$, it holds for all $\deltatil > 0$ that
    \begin{align}\label{eq:norm_comparison}
        \ee\left[\sup_{f \in \cF} \norm{f}_\nu^2 -  \frac{2}{cc'} \cdot \norm{f}_T^2 \right] \leq \deltatil^2  + \abs{\cN\left(\cF,  \frac{\sigma^2 c c' \deltatil^2}{576 \log(T)} \right)} \cdot \exp\left( - \frac{(c' \sqrt{T})^2}{72} \right) + \frac 2T.
    \end{align}
\end{theorem}
As an example, if $\cF$ is parametric in the sense that $\cN(\cF, \epsilon) \lesssim \epsilon^{-d}$ for some $d$ (e.g. when $\vc(\cF) \leq d$), \Cref{thm:small_ball} implies that the decoupled `population' norm of any data-dependent $\fhat$ can be bounded in expectation by a multiple of the empirical norm, up to a $\Otil\left( T^{-1} \right)$ term, as long as $T = \Omegatil\left( d \sigma^{-1}\log\left( \frac{1}{\sigma} \right) \right)$.  In contradistinction, applying \Cref{lem:uniform_deviations} directly only allows control up to an additive $\Otil\left( T^{-1/2} \right)$ error, which results in much weaker bounds.

By the above reasoning, \Cref{thm:small_ball} is a major improvement over uniform deviations style bounds, but one might naturally wonder how limiting \eqref{eq:smallball} is as an assumption.  Note that the small-ball condition reflects the interaction between the measure $\mu$ and the function class $\cF$, and is motivated by that in \citet{mendelson2015learning}.  Unlike in that earlier work, however, simple hypercontractivity arguments coupled with the lemma of Paley-Zygmund do not suffice to ensure \eqref{eq:smallball} due to the fact that the small ball probability must be much smaller (certainly bounded by $O(\sigma)$) than is required in the standard small-ball argument.  Two cases where \eqref{eq:smallball} does hold, however, may illuminate the generality of \Cref{thm:small_ball}.  First, if $\cF: \cX \to \left\{ \pm1, 0 \right\}$ is a class of differences of binary-valued functions, and $\gamma = \inf_{f \in \cF} \mu\left(f(Z) = 0  \right)$\footnote{If $\cF = \cG - \cG$, then $\gamma$ is the minimal probability that any two functions agree and thus $\gamma > 0$ amounts to a gap condition on $\cG$ that intuitively characterizes the instance-dependent difficulty of identifying a given $g \in \cG$ from data.}, then as long as $\sigma = \Omega(\gamma)$, it is immediate that \eqref{eq:smallball} holds with $c = \frac{\sigma}{4}$ and $c' = \Omega(1)$.  Second, if $\cX \subset \rr^d$, $\mu$ has bounded density with respect to the Lebesgue measure and $\cF$ satisfies the condition that $f(Z)$ has bounded density with respect to Lebesgue, as is common if $f$ is continuous \citep{rudin1976principles}, then taking $c = \Theta(\sigma)$ and $c' = \Omega(1)$ suffices to ensure \eqref{eq:smallball}.  We remark that an extension of this example is implied by the \emph{trajectory small-ball} condition of \citet{tu2022learning}, which was used to prove similar norm comparison guarantees for linear classes.  In \citet[Section 4.1]{tu2022learning} the authors provide many examples of data sequences satisfying this condition.  Thus, \Cref{thm:small_ball} can be seen as a nonlinear generalization of the linear norm comparison results for sequential data found in earlier work \citep{simchowitz2018learning,tu2022learning}.

In both of the above cases, we note that the pre-factor $2/(cc')$ in front of the expected empirical norm contains a polynomial dependence on $\sigma^{-1}$ which is otherwise absent from \eqref{eq:norm_comparison}; we observe that this dependence is generic.  Indeed, because $f$ is assumed bounded, if $c \gg \sigma$, then, deterministically, $\abs{f(Z)} \lesssim 1 \ll \sqrt{\frac{2c}{\sigma}} \cdot \norm{f}_\mu$ for all $\norm{f}_\mu \gtrsim \sqrt{\sigma}$.   Thus, in any nontrivial application, the prefactor in \eqref{eq:norm_comparison} should be understood to scale polynomially in $\sigma^{-1}$.

Finally, we remark that \Cref{thm:small_ball} intuitively captures a `reverse inequality' for smoothed data under the small-ball condition \eqref{eq:smallball}.  Indeed, smoothness of a measure $p$ implies that for any $f$, we may bound $\norm{f}_p \lesssim \norm{f}_\mu$, uniformly over functions $f$.  Because \Cref{thm:small_ball} applies to \emph{arbitrary} smooth measures, the conclusion yields the reverse inequality, suggesting that $\norm{f}_\mu \lesssim \norm{f}_T$ as long as $\cF$ is not too complicated.  This reverse bound is a consequence of the fact that \eqref{eq:smallball} is stronger than standard small ball assumptions in that the small ball probability must tend toward zero with $\sigma$ as opposed to remaining constant, which suffices in the easier, iid setting \citep{koltchinskii2015bounding,mendelson2015learning}.

\subsection{Proof of Theorem \ref{thm:small_ball}}\label{app:small_ball_main_proof}
We now prove \Cref{thm:small_ball}.  The proof begins by applying an argument similar to \citet{mendelson2015learning}, which applies to independent data.  This argument uses the small ball assumption \eqref{eq:smallball} to reduce the proof to controlling the uniform deviations of a function class related to $\cF$ in high probability.  We accomplish this high probability control through a discretization argument and reliance on the smoothness of the data.

Fix $f \in \cF$  and let $\nu$ be $\sigma$-smooth with respect to $\mu$.  Fix $\deltatil > 0$ and compute pointwise for $c,c'$ as in \eqref{eq:smallball},
\begin{align}
    \norm{f}_\nu^2 &\leq \deltatil^2 + \norm{f}_\nu^2 \cdot \bbI\left[\norm{f}_\nu^2 \geq \deltatil^2  \right]  \\
    &\leq \deltatil^2 + \norm{f}_\nu^2 \cdot \bbI\left[ \inf_{\substack{f \in \cF \\ \norm{f}_\nu \geq \deltatil }} \frac 1T \cdot \sum_{t = 1}^T \bbI\left[ \abs{f(X_t)} \geq \sqrt  c \cdot \norm{f}_{\nu} \right] \geq \frac {c'}2 \right] \\
    &\quad + \norm{f}_\nu^2 \cdot \bbI\left[ \inf_{\substack{f \in \cF \\ \norm{f}_\nu \geq \deltatil }} \frac 1T \cdot \sum_{t = 1}^T \bbI\left[ \abs{f(X_t)} \geq \sqrt c \cdot \norm{f}_{\nu} \right] < \frac {c'}2 \right].
\end{align}
For the second term above, we note that
\begin{align}
    \norm{f}_\nu^2 \cdot \bbI\left[ \inf_{\substack{f \in \cF \\ \norm{f}_\nu \geq \deltatil }} \frac 1T \cdot \sum_{t = 1}^T \bbI\left[ \abs{f(X_t)} \geq \sqrt c \cdot \norm{f}_{\nu} \right] \geq \frac {c'}2 \right] \leq \frac{2}{cc'} \cdot \norm{f}_T^2.
\end{align}
Rearranging and taking expectations, we see that
\begin{align}
    \ee\left[ \sup_{f \in \cF} \norm{f}_\nu^2 - \frac{2}{cc'}\cdot \norm{f}_T^2 \right] \leq \deltatil^2  + \pp\left(  \inf_{\substack{f \in \cF \\ \norm{f}_\nu \geq \deltatil }} \frac 1T \cdot \sum_{t = 1}^T \bbI\left[ \abs{f(X_t)} \geq \sqrt c \cdot \norm{f}_{\nu} \right] < \frac {c'}2 \right).
\end{align}
Thus we must bound the final term above.  To do this, we compute
\begin{align}\label{eq:smallball_proof_1a}
    \inf_{\substack{f \in \cF \\ \norm{f}_\nu \geq \deltatil }} \frac 1T \cdot \sum_{t = 1}^T \bbI\left[ \abs{f(X_t)} \geq \sqrt c \cdot \norm{f}_{\nu} \right] &\geq \inf_{\substack{f \in \cF \\ \norm{f}_\nu \geq \deltatil }} \sum_{t=  1}^T \pp_{t-1}\left( \abs{f(X_t)} \geq 2 \sqrt c \cdot \norm{f}_\nu \right) \\
    &- \sup_{\substack{f \in \cF \\ \norm{f}_\nu \geq \deltatil }} \frac 1T \cdot \sum_{t = 1}^T \pp_{t-1}\left( \abs{f(X_t)} \geq 2\sqrt c \cdot \norm{f}_\nu \right) - \bbI\left[ \abs{f(X_t)} \geq \sqrt c \cdot \norm{f}_\nu \right].
\end{align}
By \Cref{lem:mu_smallball}, and the assumption that \eqref{eq:smallball} applies, it holds that for any $t$,
\begin{align}
    \inf_{\substack{f \in \cF \\ \norm{f}_{\nu} \geq \deltatil}} \pp_{t-1}\left(\abs{f(X_t)} \geq 2\sqrt c \cdot \norm{f}_{\nu}  \right) \geq c'
\end{align}
and so the first term in \eqref{eq:smallball_proof_1a} is at least $c'$.  Thus we focus on bounding the second term in \eqref{eq:smallball_proof_1a}.  To do this, define the function
\begin{align}
    \phi_c(u) = 
    \begin{cases}
        0 & \abs{u} \leq \sqrt c \\
        u / \sqrt c - 1 & \sqrt c \leq \abs{u} \leq 2\sqrt c \\
        1 & \abs{u} \geq 2\sqrt c
    \end{cases}
\end{align}
and observe that
\begin{align}
    \sup_{\substack{f \in \cF \\ \norm{f}_{\nu} \geq \deltatil}} &\frac 1T \cdot \sum_{t = 1}^T \pp_{t-1}\left( \abs{f(X_t)} \geq 2\sqrt c \cdot \norm{f}_{\nu}  \right) - \bbI\left[ \abs{f(X_t)} \geq \sqrt c \cdot \norm{f}_{\nu} \right] \\
    &\leq \sup_{\substack{f \in \cF \\ \norm{f}_{\nu} \geq \deltatil}} \frac 1T \cdot \sum_{t = 1}^T \ee_{t-1}\left[ \phi_c\left( \frac{f(X_t)}{\norm{f}_{\nu}} \right) \right] - \phi_c\left( \frac{f(X_t)}{\norm{f}_{\nu}} \right). \label{eq:contraction_smallball}
\end{align}

In order to bound this last expression with high probability\footnote{In \citet{mendelson2015learning}, the conclusion of the proof is simpler, as concentration and contraction can directly be applied to \eqref{eq:contraction_smallball}.  Unfortunately, neither concentration nor contraction directly apply in the smoothed data setting, requiring alternative techniques.}, we will apply \Cref{lem:uniform_deviations_smooth_hp} to the function class
\begin{align}\label{eq:Gdelta_def}
    \cG_{\deltatil} = \left\{ \phi_c\left( \frac{f}{\norm{f}_{\nu}} \right) \bigg| f \in \cF \text{ and } \norm{f}_{\nu} \geq \deltatil \right\}.
\end{align}
Observing that $\cG_{\deltatil}$ is bounded, we may apply \Cref{lem:uniform_deviations_smooth_hp} to see that
\begin{align}
    \pp\left( \sup_{g \in \cG_{\deltatil}} \frac 1T \cdot \sum_{t = 1}^T \ee_{t-1}[g(X_t)] - g(X_t) > v \right) \leq \abs{\cN(\cG_{\deltatil}, \epsilon)} \cdot \exp\left( - \frac{T v^2}{18} \right) + T e^{- \sigma k} + \delta,
\end{align}
as long as $\sigma k \geq 1$ and 
\begin{align}
    v \geq 6 k \epsilon + 6 \cdot \sqrt{\frac kT\left( \log\left( \frac{\cN(\cG, \epsilon)}{\delta} \right) \right)}.
\end{align}
Taking
\begin{align}
    \delta = \frac 1T, \qquad k = \frac{3 \log(T)}{\sigma}, \qquad \epsilon = \frac{c'}{24 k}, \qquad \text{and} \qquad v = \frac{c'}{2},
\end{align}
we see that whenever
\begin{align}
    \frac{T}{\log^2(T) \cdot \log \cN\left( \cG, \frac{\sigma c'}{72 \log(T)} \right)} \geq \frac{576}{\sigma},
\end{align}
it holds that
\begin{align}
    \pp\left( \sup_{g \in \cG_{\deltatil}} \frac 1T \cdot \sum_{t = 1}^T \ee_{t-1}[g(X_t)] - g(X_t) > \frac {c'}2 \right) \leq \abs{\cN\left(\cG_{\deltatil}, \frac{\sigma c'}{72 \log(T)}\right)} \cdot \exp\left( - \frac{(\sqrt{T} c')^2}{72} \right) + \frac 2T
\end{align}
By \Cref{lem:Gdelta_cover_bound}, then, it holds that
\begin{align}
    \pp\left( \sup_{g \in \cG_{\deltatil}} \frac 1T \cdot \sum_{t = 1}^T \ee_{t-1}[g(X_t)] - g(X_t) > \frac {c'}2 \right) \leq \abs{\cN\left(\cF,  \frac{\sigma^2 c c' \deltatil^2}{576 \log(T)} \right)} \cdot \exp\left( - \frac{(\sqrt{T} c')^2}{72}\right) + \frac 2T.
\end{align}
Thus,
\begin{align}
    \pp&\left(\sup_{\substack{f \in \cF \\ \norm{f}_{\nu} \geq \deltatil}} \frac 1T \cdot \sum_{t = 1}^T \pp\left( \abs{f(X_t)} \geq 2\sqrt c \cdot \norm{f}_{\nu}  \right) - \bbI\left[ \abs{f(X_t)} \geq \sqrt c \cdot \norm{f}_{\nu} \right] > \frac{c'}{2}  \right) \\
    &\leq \abs{\cN\left(\cF,  \frac{\sigma^2 c c' \deltatil^2}{576 \log(T)} \right)} \cdot \exp\left( - \frac{(\sqrt{T} c')^2}{72}\right) + \frac 2T.
\end{align}
Plugging this into \eqref{eq:smallball_proof_1a}, we see that
\begin{align}
    \pp\left(\inf_{\substack{f \in \cF \\ \norm{f}_{\nu} \geq \deltatil}} \frac 1T \cdot \sum_{t = 1}^T \bbI\left[ \abs{f(X_t)} \leq\sqrt  c \cdot \norm{f}_{\nu} \right]  \leq \frac{c'}{2}  \right) \leq \abs{\cN\left(\cF,  \frac{\sigma^2 c c' \deltatil^2}{576 \log(T)} \right)} \cdot \exp\left( - \frac{(c' \sqrt{T})^2}{72} \right) + \frac 2T.
\end{align}
The result follows. \iftoggle{colt}{\jmlrQED}{\qed}

\subsection{Auxiliary Lemmata}\label{app:small_ball_lemmata}
In this section we prove a number of auxiliary results that are used in the proof of \Cref{thm:small_ball}.  We begin with a lemma that ensures that \eqref{eq:smallball} implies a small-ball like condition for all smooth measures.
\begin{lemma}\label{lem:mu_smallball}
    Suppose that $\cF: \cX \to [-1,1]$ is a function class and $\mu \in \Delta(\cX)$.  Suppose that \eqref{eq:smallball} holds for $c, c' > 0$.  Then it holds for any $\nu, p \in \Delta(\cX)$ such that $p$ and $\nu$ are $\sigma$-smooth with respect to $\mu$ that
    \begin{align}
        \inf_{f \in \cF} \nu\left( \abs{f(Z)} \geq 2 \sqrt c \cdot \norm{f}_{p} \right) \geq c'.
    \end{align}
\end{lemma}
\begin{proof}
    Note that by definition of the Radon-Nikodym derivative, it holds that $\norm{f}_{p} \leq \sigma^{-1/2} \cdot \norm{f}_\mu$.  Thus we may compute that
    \begin{align}
        \nu\left( \abs{f(Z)} < 2 \sqrt c \cdot \norm{f}_{p} \right)  \leq \frac{1}{\sigma} \cdot \mu\left(  \abs{f(Z)} < 2 \sqrt{\frac{c}{\sigma}} \cdot \norm{f}_{\mu} \right) \leq \frac{1}{\sigma}  \sigma (1 - c') = 1 - c',
    \end{align}
    where the second inequality follows by the definition of smoothness and the last inequality follows by \eqref{eq:smallball}.  The result follows.
\end{proof}
We now prove a uniform deviations result akin to \Cref{lem:uniform_deviations} below, except that it holds in high probability instead of in expectation.  For this to work, we modify the notion of complexity to covering number from Rademacher complexity.
\begin{lemma}\label{lem:uniform_deviations_smooth_hp}
    Let $\cG: \cX \to [-1,1]$ be a function class, $\mu \in \Delta(\cX)$ a measure, and suppose that $X_1, \dots, X_T$ are $\sigma$-smooth with respect to $\mu$.  Fix $k \in \bbN$ and suppose that $\cN(\cG, \epsilon)$ denote the covering number of $\cG$ at scale $\epsilon$ with respect to $\norm{\cdot} = \norm{\cdot}_\mu$.  Suppose that $k, \sigma, \epsilon > 0$ such that $k \sigma \geq 1$ and
    \begin{align}
        v > 6 k \epsilon + 6 \cdot \sqrt{\frac kT\left( \log\left( \frac{\cN(\cG, \epsilon)}{\delta} \right) \right)}.
    \end{align}
    Then it holds that
    \begin{align}
        \pp\left( \sup_{g \in \cG} \frac 1T \sum_{t = 1}^T \ee_{t-1}[g(X_t)] - g(X_T) > v \right) \leq \abs{\cN(\cG, \epsilon)} \cdot \exp\left( - \frac{T v^2}{18} \right) + T e^{- \sigma k} + \delta.
    \end{align}
\end{lemma}
\begin{proof}
    We prove this result by discretizing to the cover and then applying a standard concentration bound to bounded martingale difference sequences.  To do this, let $\pi : \cG \to \cN(\cG, \epsilon)$ denote projection onto the cover.  Then by a union bound, we see that for any $v > 0$,
    \begin{align}
        \pp\left( \sup_{g \in \cG} \frac 1T \cdot \sum_{t = 1}^T \ee_{t-1}[g(X_t)] - g(X_T) > v \right) &\leq \pp\left( \max_{g \in \cN(\cG, \epsilon)} \frac 1T \cdot  \sum_{t = 1}^T \ee_{t-1}[g(X_t)] - g(X_T) > \frac{v}{3}   \right) \\
        &\quad + \pp\left( \sup_{g \in \cG} \frac 1T \cdot \sum_{t = 1}^T \abs{\ee_{t-1}[g(X_t)] - \ee_{t-1}[\pi(g(X_t))]} > \frac{v}{3} \right) \\
        &\quad + \pp\left( \sup_{g \in \cG}\frac 1T \cdot  \sum_{t = 1}^T \abs{g(X_t) - \pi(g(X_t))} > \frac{v}{3} \right). \label{eq:uniform_deviations_smooth_hp_1}
    \end{align}
    For the first term, note that by Azumas's inequaltiy \citep{azuma1967weighted}, it holds for any fixed $g \in \cG$ that
    \begin{align}
        \pp\left( \sum_{t = 1}^T \ee_{t-1}\left[ g(X_t) \right] - g(X_t) > u \right) \leq \exp\left( - \frac{u^2}{2 T} \right).
    \end{align}
    Thus by a union bound, it holds that
    \begin{align}
        \pp\left( \max_{g \in \cN(\cG, \epsilon)} \frac 1T \cdot  \sum_{t = 1}^T \ee_{t-1}[g(X_t)] - g(X_T) > \frac v3   \right) \leq \abs{\cN(\cG, \epsilon)} \cdot \exp\left( - \frac{T v^2}{18} \right).
    \end{align}

    For the second term, we see that by smoothness, for all $t$,
    \begin{align}
        \abs{\ee_{t-1}[g(X_t)] - \ee_{t-1}[\pi(g(X_t))]} &\leq \sigma^{-1/2} \cdot \norm{g - \pi(g)}_\mu \leq \frac \epsilon {\sqrt \sigma}.
    \end{align}
    Thus the second term vanishes as long as $v \geq 3 \epsilon / \sqrt{\sigma}$.

    Finally, for the third term, let $\cE$ denote the event from \Cref{lem:coupling} and observe that
    \begin{align}
        \pp\left( \sup_{g \in \cG} \frac 1T \cdot \sum_{T = 1}^T \abs{g(X_t) - \pi \circ g(X_t)} > \frac v3 \right) &= \pp\left( \left\{ \sup_{g \in \cG}\frac 1T \cdot  \sum_{t = 1}^T \abs{g(X_t) - \pi \circ g(X_t)} > \frac v3 \right\}  \cap \cE \right) \\
        &\quad + \pp\left( \left\{ \sup_{g \in \cG}\frac 1T \cdot  \sum_{t = 1}^T \abs{g(X_t) - \pi \circ g(X_t)} > \frac v3 \right\}  \cap \cE^c \right) \\
        &\leq \pp\left( \sup_{g \in \cG}\frac 1T \cdot  \sum_{t = 1}^T \sum_{j = 1}^k \abs{g(Z_{t,j}) - \pi\circ g(Z_{t,j})} > \frac v3  \right) \\
        &\quad + T e^{- \sigma k} \\
        &\leq \pp\left( \sup_{g \in \cG}\frac 1{kT} \cdot  \sum_{t = 1}^T \sum_{j = 1}^k \abs{g(Z_{t,j}) - \pi\circ g(Z_{t,j})} > \frac v{3k}  \right) + T e^{-\sigma k}.
    \end{align}
    Noting now that the $Z_{t,j}$ are independent and identically distributed, and applying standard high probability uniform concentration (e.g., \citet[Theorem 4.10]{wainwright2019high}), we have that with probability at least $1 - \delta$,
    \begin{align}
        \sup_{g \in \cG}\frac 1{kT} \cdot  \sum_{t = 1}^T \sum_{j = 1}^k \abs{g(Z_{t,j}) - \pi\circ g(Z_{t,j})} &\leq \epsilon + 2 \cdot \frac{\rad_{kT}(\cG)}{kT} +  2 \cdot \sqrt{\frac{\log\left( \frac 1\delta \right)}{kT}} \\
        &\leq 2 \epsilon + 2 \cdot \sqrt{\frac{\log\left( \cN(\cG, \epsilon) \right) + \log\left( \frac 1\delta \right)}{k T}},
    \end{align}
    where the second inequality follows by standard bounds on Rademacher complexity by covering numbers (e.g. \citet[Corollary 5.25]{van2014probability}).  Thus, as long as
    \begin{align}
        v > 6 k \epsilon + 6 \cdot \sqrt{\frac kT\left( \log\left( \frac{\cN(\cG, \epsilon)}{\delta} \right) \right)},
    \end{align}
    the third term is bounded by $\delta$.  The result follows.
\end{proof}
Finally, we prove a result akin to contraction, showing that if $\cG_{\deltatil}$ is as in \eqref{eq:Gdelta_def}, then the covering number of $\cG_{\deltatil}$ is upper bounded by the covering number of $\cF$.
\begin{lemma}\label{lem:Gdelta_cover_bound}
    Let $\cF: \cX \to [-1,1]$ be a function class with $\norm{\cdot}_{\nu}$ and $\norm{\cdot}_{kT}$ as in \Cref{def:covering_number}.  For $\deltatil > 0$, let $\cG_{\deltatil}$ be as in \eqref{eq:Gdelta_def}.  Let $\cN(\cF, \epsilon)$ denote the covering number of $\cF$ at scale $\epsilon$ with respect to $\norm{\cdot}_\mu$.  Then for any $1 \geq c, \deltatil > 0$, it holds that
    \begin{align}
        \cN\left( \cG_{\deltatil}, \epsilon \right) \leq \cN\left( \cF, \frac{c \sigma \deltatil^2}{8} \cdot \epsilon \right).
    \end{align}
\end{lemma}
\begin{proof}
    We compute for any $X \in \cX$, and any $f,f' \in \cF$,
    \begin{align}
        \abs{\frac{f(X)}{\norm{f}_{\nu}} - \frac{f'(X)}{\norm{f'}_{\nu}}} &\leq \frac{\abs{f(X) - f'(X)}}{\norm{f}_{\nu}} + \abs{f'(X)} \cdot \abs{\frac{1}{\norm{f}_{\nu}} - \frac{1}{\norm{f'}_{\nu}}} \\
        &\leq \frac{\abs{f(X) - f'(X)}}{\norm{f}_{\nu}} + \frac{\norm{f - f'}_{\nu}}{\norm{f}_{\nu} \cdot \norm{f'}_{\nu}},
    \end{align}
    where the second inequality follows by the boundedness of $\cF$ and the triangle inequality.  If $\norm{f}_{\nu} \geq \deltatil$, then, we have that
    \begin{align}
        \abs{\frac{f(X)}{\norm{f}_{\nu}} - \frac{f'(X)}{\norm{f'}_{\nu}}} \leq \frac 1{\deltatil} \cdot \abs{f(X) - f'(X)} + \frac{1}{\deltatil^2} \cdot \norm{f -f'}_{\nu}.
    \end{align}
    Noting that $\phi_c$ is $\frac{1}{c}$-Lipschitz, we see that
    \begin{align}
        \norm{\phi_c\left( \frac{f}{\norm{f}_{\nu}} \right) - \phi_c\left( \frac{f'}{\norm{f'}_{\nu}} \right)}_\mu &\leq \frac 1c \cdot \left( \frac 1{\deltatil} \cdot \norm{f -f'}_\mu + \frac{1}{\deltatil^2} \cdot \norm{f -f'}_{\nu} \right) \\
        &\leq \frac{2}{c \sigma \deltatil^2} \cdot \norm{f - f'}_\mu,
    \end{align}
    where we used the fact that $\deltatil \leq 1$.  The result follows immediately.
\end{proof}

\end{document}